\documentclass{article} 
\usepackage[final]{colm2026_conference}

\usepackage{microtype}
\usepackage{hyperref}
\usepackage{url}
\usepackage{booktabs}
\usepackage{graphicx}
\usepackage{subcaption}
\usepackage{booktabs}
\usepackage{amsmath}
\usepackage{tabularx}
\usepackage{listings}
\usepackage{makecell}
\usepackage{arydshln}
\usepackage{float}
\usepackage{xcolor}
\usepackage{listings}
\usepackage[T1]{fontenc}
\usepackage[most]{tcolorbox}
\usepackage{arydshln}
\newtcblisting{promptbox}[1][]{
    colback=gray!5,
    colframe=gray!50,
    boxrule=0.5pt,
    arc=2pt,
    left=5pt, right=5pt, top=5pt, bottom=5pt,
    listing only,
    listing options={
        basicstyle=\ttfamily\small,
        breaklines=true,
        breakatwhitespace=true,
        columns=fullflexible,
        keepspaces=true
    },
    title={\textbf{System Prompt}},
    enhanced,
    breakable,
    before skip=\medskipamount,
    after skip=\medskipamount,
    #1
}

\usepackage{lineno}

\definecolor{darkblue}{rgb}{0, 0, 0.5}
\hypersetup{colorlinks=true, citecolor=darkblue, linkcolor=darkblue, urlcolor=darkblue}

\title{Agent Psychometrics:
Task-Level Performance Prediction \\ in Agentic Coding Benchmarks}

\author{
\parbox{\textwidth}{\centering
Chris Ge$^{1,2}$\thanks{Equal contribution; order randomized.\quad$^{\dagger}$Primary advising.} \quad
Daria Kryvosheieva$^{1,2}$\footnotemark[1] \quad
Daniel Fried$^{3}$ \\
Uzay Girit$^{2\,\dagger}$ \quad
Kaivalya Hariharan$^{2\,\dagger}$
} \\[10pt]
\parbox{\textwidth}{\centering \normalfont
$^1$MIT \quad $^2$Fulcrum \quad $^3$Carnegie Mellon University
}}

\begin{document}

\ifcolmsubmission
\linenumbers
\fi

\maketitle

\begin{abstract}
As the focus in LLM-based coding shifts from static single-step code generation to multi-step agentic interaction with tools and environments, understanding which tasks will challenge agents and why becomes increasingly difficult. This is compounded by current practice: agent performance is typically measured by aggregate pass rates on benchmarks, but single-number metrics obscure the diversity of tasks within a benchmark. We present a framework for predicting success or failure on individual tasks tailored to the agentic coding regime. Our approach augments Item Response Theory (IRT) with rich features extracted from tasks, including issue statements, repository contexts, solutions, and test cases, and introduces a novel decomposition of agent ability into LLM and scaffold ability components. This parameterization enables us to aggregate evaluation data across heterogeneous leaderboards and accurately predict task-level performance for unseen benchmarks, as well as unseen LLM-scaffold combinations. Our methods have practical utility for benchmark designers, who can better calibrate the difficulty of their new tasks without running computationally expensive agent evaluations.\footnote{Correspondence to \texttt{\{cge7, daria\_k\}@mit.edu}, \texttt{kaivu@fulcrum.inc}. Code available at \url{https://github.com/dariakryvosheieva/agent-psychometrics}.}

\end{abstract}

\section{Introduction}
\label{sec:intro}

As language models become capable of solving longer-horizon \citep{kwa2025measuring}, more complex tasks, evaluations have shifted from simple question-and-answer tasks to multi-turn assessments of a model's capability as an \textit{agent}. Nowhere is this transition more pronounced than in software benchmarking: newer benchmarks like SWE-bench Verified \citep{openai-2024} and Terminal-Bench 2.0 \citep{merrill2026terminalbench}, which test how coding agents iterate with execution feedback from tool calls, are replacing static single-step code generation benchmarks like HumanEval \citep{chen2021evaluating} and MBPP \citep{austin2021program}.

But the multi-step nature of coding agent benchmarking creates many challenges for benchmark designers. Agentic evaluations are complicated by the interaction of the agent with the environment: test cases often test for properties of components of the codebase that are not mentioned in the problem statement \citep{openai-2024} or allow underspecified submissions to pass \citep{wang2025solved}, and agentic tasks often have multiple valid solution paths and edge cases that are difficult to anticipate when designing the evaluation rubric. Indeed, such
validity issues are pervasive enough to mis-estimate agent performance by
up to 100\% across widely used benchmarks \citep{zhu2026establishing}. Worsening the challenge, agentic tasks are often heterogeneous: agents can fail on different tasks in the same benchmark for different reasons, which reporting a single overall solve rate, as is common practice \citep{anthropic-2026, openai-2026}, fails to capture. Both agent developers and benchmark designers would benefit from a task-level understanding of where agents fail, for improving agent capabilities and designing more discriminative tasks \citep{liu2025empirical}.

This kind of task-level understanding requires \textit{evaluation data}: per-agent, per-task records of which tasks each agent solved. Obtaining such data for every agent and task of interest is not always realistic: the average agent evaluation run on SWE-bench Verified costs $\$350-\$500$ \citep{jimenez2024swebench}, and the cases where task-level understanding matters most involve agents or tasks for which no evaluation data exists at all, such as a newly released agent that has not been tested on any benchmark, or a draft benchmark whose tasks no agent has attempted. This motivates the central question of this work: \emph{how can we efficiently predict agent performance at the task level in agentic coding benchmarks?}

Our approach extends Item Response Theory (IRT), a method from psychometrics for modeling the interaction between test-takers and exam problems \citep{baker2001basics} that takes evaluation data as its foundational input. In IRT applied to model evaluations, agents are test-takers with latent ability scores, and benchmark tasks are exam problems with latent difficulty scores \citep{hofmann2025fluid}; the success probability of any agent on any task can be predicted as a function of the agent's ability and the task's difficulty (see Section~\ref{sec:irt}). Standard IRT methods are limited by treating each task and each agent as an ID: they can only score agents and tasks that already appear in the evaluation data, and offer no insight into what aspects of the task explain its difficulty.

We address these limitations by parameterizing IRT abilities and difficulties as functions of agent and task \textit{features}, enabling task-level performance prediction even when no evaluation data is available for a given agent or task. This builds on the work of \citet{chen2025learning} and \citet{truong2025reliable}, who use linear models to estimate IRT task difficulties from task embeddings and thereby predict performance on question-and-answer tasks from the problem statement alone. We incorporate their task difficulty predictor as part of our success predictor, but with different inputs specialized for the agentic coding regime, where success derives from the complex interaction of properties of the agent, task, and environment.

\begin{figure}[t]
    \centering
    \includegraphics[width=\textwidth]{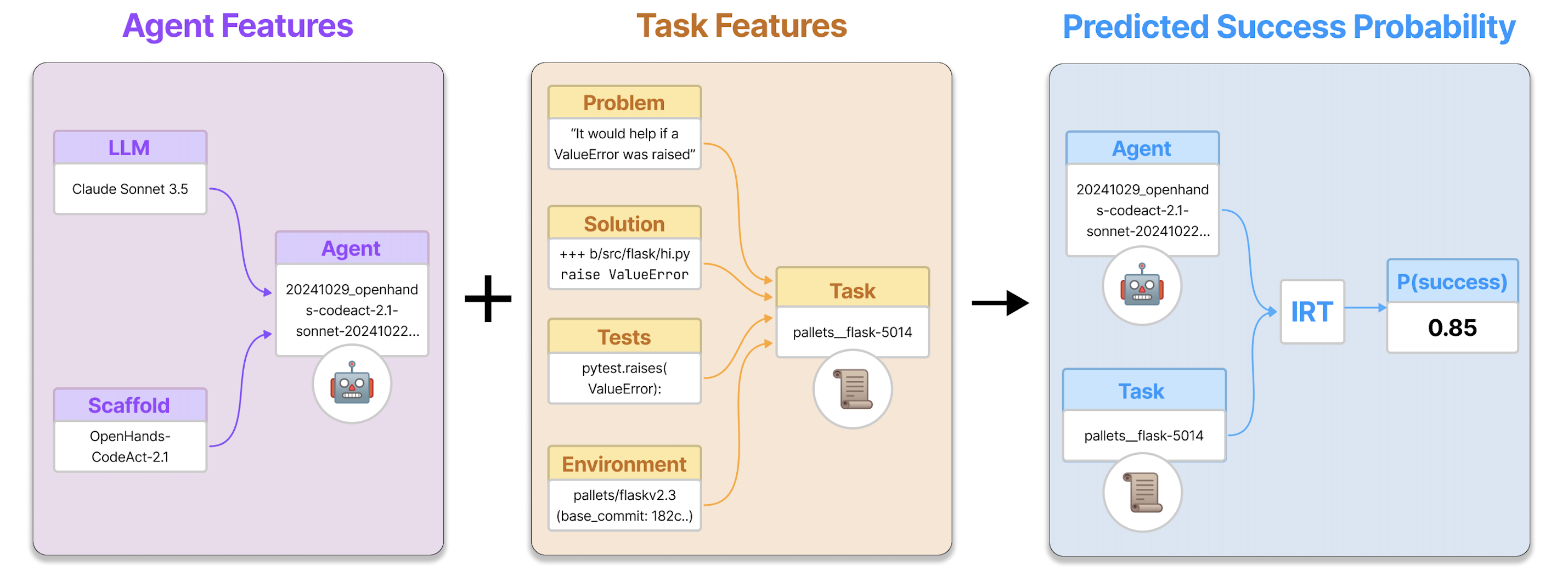}
    \caption{\textbf{Agent and task features predicting success probability.} On the agent side, the features are two categorical variables, the LLM and the scaffold; on the task side, they are vectors extracted from rich task artifacts. From these we estimate the agent's \textit{ability} and the task's \textit{difficulty}, then apply the logistic model from IRT \citep{baker2001basics} to predict the probability that the agent succeeds on the task.}
    \label{fig:overview_figure}
\end{figure}

We find that features specific to the agentic coding setting offer enhanced predictive capabilities: for agents, we use the underlying LLM and scaffold, and for tasks, dense (non-one-hot) feature vectors extracted from the task's test cases, solution patch, and repository state, in addition to the problem statement. We demonstrate that using these features, we can meaningfully predict success probability for held-out tasks and LLM-scaffold combinations. This ability to generalize beyond observed evaluations has practical value for the developers of both benchmarks and agents: benchmark designers can estimate the difficulty of tasks they have not yet evaluated agents on, including in continuously evolving benchmarks such as SWE-bench Live \citep{zhang2025swebenchgoeslive}, and agent developers can evaluate a new LLM-scaffold pair on a small, adaptively chosen subset of tasks, as we demonstrate in Section~\ref{sec:new-agents-benchmark}.

A summary of our major contributions is as follows: 
\begin{enumerate}
\item \textbf{Difficulty prediction from agentic task features:}
We extend task difficulty prediction from  \citet{truong2025reliable} to agentic coding benchmarks by learning predictors from static task features (issue text, gold patch, test patch, and repository state). 

\item \textbf{LLM-scaffold ability decomposition for multi-benchmark IRT:} We learn an IRT ability component for each LLM and each scaffold that appears in an agent individually, allowing us to aggregate  benchmarks whose evaluation data share few overlapping agents. We validate the learned LLM abilities by comparison to a fixed-scaffold setting, and further demonstrate that the multi-benchmark IRT model can be used to predict performance for held-out benchmarks and held-out agents whose LLM and scaffold are individually present in the agent evaluation data. 
\end{enumerate}

\section{Background and Related Work}

\subsection{Background}
\label{sec:background}

\subsubsection{Agentic Coding Benchmarks}

\label{sec:agentic-coding-benchmarks}

In contrast to question-and-answer benchmarks, agentic benchmarks require the LLM to dynamically call tools and explore the environment, eventually submitting a final solution, which is validated via unit tests or observations of the environment's final state. Agentic benchmarks are especially popular in the coding domain because many programming or software engineering tasks naturally involve interacting with entire codebases. To function as an agent, an LLM needs to be augmented with a \textit{scaffold}, or a framework comprising tools, system prompts, and often a retrieval system \citep{grace2026demystifying}.

In our experiments, we consider four  human-verified agentic coding benchmarks, covering a wide range of coding skills:
\begin{itemize}
\item \textbf{SWE-bench Verified} \citep{openai-2024} is a human-verified subset of tasks from SWE-bench \citep{jimenez2024swebench}, a popular general-purpose software engineering benchmark. The tasks involve solving real-world GitHub issues from Python repositories. \item \textbf{SWE-bench Pro} \citep{deng2025swebenchproaiagents} is more challenging than SWE-bench Verified and covers a wider variety of repositories while using a similar task format. \item \textbf{Terminal-Bench 2.0} \citep{merrill2026terminalbench} is a challenging benchmark focused on computer terminal environments. \item \textbf{GSO} \citep{shetty2025gsochallengingsoftwareoptimization} is a very challenging benchmark focused on software performance optimization. To succeed, an agent must optimize a program to reach at least 95\% of the speed of the ground-truth (expert-written) optimization while passing correctness tests.
\end{itemize}

For each dataset, we use publicly available agent evaluation data\footnote{Per-dataset sources are listed in our GitHub repository.}, showing a detailed breakdown of how each agent performed on each task. For Terminal-Bench 2.0, since agents had multiple attempts at each task, we take the majority ($\ge 50\%$) outcome as the binary response for each agent-task pair (see Appendix \ref{app:binomial} for experiments using all attempts).

\subsubsection{Item Response Theory (IRT)}
\label{sec:irt}

IRT is a technique from classical psychometrics for analyzing the interaction between test takers and test problems \citep{Lord1968, baker2001basics}. In the simplest IRT model---the one-dimensional one-parameter logistic (1D 1PL) model, also known as the Rasch model---each test-taker $i$ has a latent ability parameter $\theta_i$ and each problem $j$ has a latent difficulty parameter $\beta_j$ \citep{rasch1993probabilistic}. Together, these parameters determine the predicted probability that test-taker $i$ succeeds on problem $j$:
\begin{equation}
\label{eq:irt-prediction}
P(y_{ij} = 1 \mid \theta_i, \beta_j) = \sigma(\theta_i - \beta_j),
\end{equation}
where $y_{ij} \in \{0, 1\}$ is the observed response of test-taker $i$ on problem $j$ (1 = correct, 0 = incorrect) and $\sigma$ is the sigmoid function. In our experiments, the IRT model is fitted by maximizing the evidence lower bound (ELBO) via stochastic variational inference (SVI) with hierarchical priors. IRT has been used in standardized tests like the SAT to provide well-calibrated measures of test-taker ability \citep{petersen1982using, sat}.

\subsection{Related Work}
\label{sec:related-work}

IRT was first applied in NLP by \citet{lalor2016building}, who used the IRT ability instead of the accuracy score as a better measure of a model's capability. Later works used IRT to obtain faster evaluations of LLM ability by evaluating on a small curated subset of the benchmark data \citep{polo2024tinybenchmarks}, later extended to adaptively chosen subsets \citep{li2025adaptive, hofmann2025fluid}, but they all do so with IRT difficulty scores derived from task-level response data, which is not publicly available for many major agentic benchmarks \citep{chan2025mlebench, starace2025paperbench, wijk2025rebench, zhou2026featurebench}.

Recent studies have increasingly focused on the IRT difficulty parameter. \citet{liu2026bridge} found that IRT difficulty is linearly related to the log of the human completion time of a task,  allowing them to convert from response data to a completion time horizon estimate. 
\citet{truong2025reliable}, \citet{chen2025learning}, and \citet{zhou2026general} all  developed methods to estimate task difficulties without response data; in particular, \citet{truong2025reliable} used a linear model to predict difficulties from an embedding of the task statement. We adapt their methodology to the agentic coding setting, leveraging features of tasks and test-takers that are specific to this setting, and we compare directly to their original approach in Section~\ref{sec:new-tasks}. \citet{truong2025reliable} also estimated test-taker (LLM) ability from training compute in FLOPs, which is unavailable for many LLMs, especially proprietary ones. In contrast, our agent features, the LLM and scaffold, are more likely to be known.

\section{Methods}
\label{sec:methods}

We address IRT's generalization limitations by extending it to capture informative features of tasks and agents. This enables us to predict agent performance for new tasks without response data, as well as (more limitedly) task performance for new agents without response data.

\subsection{Task Feature Encodings}
\label{sec:task-feature-encodings}

We consider two methods of converting a task into a meaningful feature vector:
\begin{enumerate}
\item \textbf{embeddings} (feeding the task as input into an open-weight LLM and obtaining its embedding vector);
\item \textbf{LLM-as-a-judge features} (prompting an LLM to grade the task according to multiple pre-defined rubric criteria, such as how easy it is to verify that the task was done correctly or how much domain-specific knowledge the task requires).
\end{enumerate}

Both methods aim to capture aspects of a task that could contribute to its difficulty, while making use of its surrounding metadata and agentic artifacts. See Appendix \ref{app:details-task-feature-extraction} for further details on how we extract the feature vectors.

Our response prediction methods are compatible with any vector encoding of a task; in particular, we can concatenate vectors of different types, which sometimes outperforms each vector alone.

\subsection{Response Prediction Methods}
\label{sec:response-prediction-methods}

In our experiments, we introduce modifications to the standard IRT model to incorporate two kinds of features. For tasks, we use feature vectors described in Section~\ref{sec:task-feature-encodings}. For agents, in contrast, we use two categorical variables (the agent's underlying LLM and scaffold); we explain this choice in Section~\ref{sec:irt-agent} and discuss why it is fair to call these categorical variables ``features'' in Appendix~\ref{app:agent_features_one_hot}. 

\subsubsection{IRT with Task Features}
\label{sec:irt-task}

Following \citet{truong2025reliable}, we fit a linear model to predict a task's IRT difficulty parameter from its feature vector. We differ from \citet{truong2025reliable} in two ways. First, we consider a wider variety of feature vectors (LLM-as-a-judge
features and concatenated vectors in addition to embeddings). Second, whereas \citet{truong2025reliable} jointly trained the IRT ability parameters and the weights of the linear model via a maximum log likelihood objective, we use a two-stage training procedure. We start by fitting standard IRT parameters $\theta_i$ and $\beta_j$ as described in Section~\ref{sec:irt}, then freeze the resulting difficulties $\beta_j$ and fit a ridge regression to predict them from task feature vectors $f_j \in \mathbb{R}^d$:
\begin{equation}
\label{eq:ridge}
\hat{w}, \hat{b} = \arg\min_{w, b} \sum_j \bigl(\beta_j - w^\top f_j - b\bigr)^2 + \lambda \|w\|_2^2.
\end{equation}
At inference time, like \citet{truong2025reliable}, we predict the difficulty of a new task $j^\ast$ with feature vector $f_{j^\ast}$ as $\hat\beta_{j^\ast} = \hat{w}^\top f_{j^\ast} + \hat{b}$, and use the predicted difficulty $\hat\beta_{j^\ast}$ together with a known agent ability $\theta_i$ to predict success probability:
\begin{equation}
\label{eq:irt-task-prediction}
P(y_{ij^\ast} = 1 \mid \theta_i, \hat\beta_{j^\ast}) = \sigma\!\bigl(\theta_i - \hat\beta_{j^\ast}\bigr).
\end{equation}
We found our two-stage training procedure to outperform joint log likelihood maximization (see Appendix \ref{app:joint_difficulty_training}). The best regularization hyperparameter $\lambda$ is chosen to minimize 5-fold cross-validation MSE within the training set. When concatenating an embedding vector with an LLM-as-a-judge vector, we use different regularization hyperparameters for each one, since the two vectors may have very different dimensions.

\subsubsection{IRT with Agent Features (LLM and Scaffold)}
\label{sec:irt-agent}

We incorporate agent features differently from task features: extracting numerical feature vectors from agents based on text inputs, as we do for tasks, would be complicated due to the lack of standardized textual descriptions of agents and the limited availability of information on proprietary agents. Rather, \textbf{we decompose an agent's IRT ability into the abilities of its underlying LLM and scaffold}. Doing so allows us to ``stitch together'' multiple benchmark leaderboards and train multi-benchmark IRT models: although agents (LLM-scaffold combinations) rarely overlap across different leaderboards, individual LLMs and scaffolds overlap much more frequently.

We relate the two ability parameters via summation; we validate this choice of functional form in Appendix \ref{app:functional-forms}. Under the proposed IRT model, the probability that LLM $m$ combined with scaffold $s$ succeeds on task $j$ is given by
\begin{equation}
\label{eq:multi_benchmark_irt}
P(y_{msj} = 1 \mid \theta_m, \theta_s, \beta_j) = \sigma(\theta_{m} + \theta_{s} - \beta_j).
\end{equation}
This approach allows us to predict responses $y_{msj}$ for new LLM-scaffold combinations $(m, s)$, as long as the LLM  $m$ and scaffold $s$ were individually seen in the training data. If we combine this approach with task features (Section \ref{sec:irt-task}) by training the proposed IRT model according to Equation \ref{eq:multi_benchmark_irt} and then training a linear model to predict $\beta_j$ from task features, we can also predict responses on new tasks from held-out benchmarks.

We note that this approach cannot account for agents with multiple or undisclosed LLMs; we exclude such agents from the experiment and discuss one possible extension to multiple LLMs in Section \ref{sec:discussion}. At evaluation time, the approach cannot generalize to agents with unseen LLMs and/or scaffolds, which we also exclude.

\section{Experimental Structure}
\label{sec:experiments}

The broad goal of our experiments is to find task and agent features that are predictive of agent success on a task. We measure the quality of our features by using them to predict success probabilities on held-out sets of responses that are chosen to test a generalization property of our predictor. Following \citet{truong2025reliable}, our main evaluation metric is AUC-ROC, which we define in more detail in Appendix \ref{app:AUC}. 

All our experiments follow the same structure:
\begin{enumerate}
    \item Hold out a set of responses (the $y_{ij}$ booleans denoting whether or not an agent succeeded on a task; see Section \ref{sec:irt}).
    \item Train a predictor (IRT with task features, agent features, or both; see Section \ref{sec:response-prediction-methods}) on the remaining responses.
    \item Use the trained predictor to obtain success probabilities on the held-out responses.
    \item Evaluate the predicted probabilities via AUC-ROC. 
\end{enumerate}

The experiments primarily vary in what set of responses is held out. The full description of each experimental setting is provided in Table \ref{tab:experimental-settings}.

\begin{table}[h]
\small
\centering
\renewcommand{\arraystretch}{1.15}
\begin{tabularx}{\textwidth}{
    >{\raggedright\arraybackslash}p{1.53cm}
    >{\raggedright\arraybackslash}p{1.15cm}
    >{\raggedright\arraybackslash}p{1.20cm}
    >{\raggedright\arraybackslash}p{2.24cm}
    >{\raggedright\arraybackslash}p{2.17cm}
    >{\raggedright\arraybackslash}X}
\toprule
Setting & \# Bench. & Method & Held-Out & Metric & Objective \\
\midrule
New Tasks 
& Single
& IRT with \textbf{task} features
& Random subset of \textbf{tasks} (in-distribution) 
& Mean 5-fold CV AUC-ROC over tasks 
& Test whether task features explain task difficulty. \\
\hline
New Responses 
& Single \& Multi
& IRT with \textbf{agent} features
& Random subset of \textbf{responses} (in-distribution) 
& Mean 5-fold CV AUC-ROC over responses 
& Test whether LLM and scaffold abilities combine additively. \\
\hline
New Agents 
& Single
& IRT with \textbf{agent} features
& \textbf{Agents} whose LLM and scaffold were individually seen 
& Mean 5-fold CV AUC-ROC over agents with seen LLMs and scaffolds 
& Test generalization to unseen model–scaffold combinations. \\
\hline
New \linebreak Benchmarks 
& Multi
& IRT with \textbf{task \& agent} features
& Entire \textbf{benchmark} (out-of-distribution) 
& Validation AUC-ROC on held-out benchmark 
& Test out-of-distribution generalization for early-stage benchmark design. \\
\bottomrule
\end{tabularx}
\caption{\textbf{Experimental settings.} We evaluate response prediction under four settings.}
\label{tab:experimental-settings}
\end{table}

Where applicable, we compare our methods against a naive \textbf{baseline}. For held-out tasks, the baseline assigns each agent a success probability equal to its average success rate across the training set; for held-out benchmarks, we use a variant of this baseline that always predicts the LLM's success rate ignoring the scaffold; for held-out agents, the baseline predicts each task's average success rate across agents from the training set. For held-out responses, we do not use a naive baseline but instead compare our method to standard IRT trained on the training set. In all experiments, we use bootstrapping to verify that our IRT-based predictors significantly beat the baseline. Because each experiment involves multiple benchmark and method comparisons, we apply a Holm-Bonferroni correction within each experiment; every comparison that is significant at the 0.05 level remains significant after the correction. The descriptions and results of the bootstrapping significance tests are outlined in Appendix \ref{app:bootstrap}.

We also compare every method to an \textbf{oracle} IRT, trained on all the data including the held-out responses and serving as an upper bound on the performance of IRT-based methods. We train our predictors per-benchmark in the \textit{New Tasks} and \textit{New Agents} settings; multi-benchmark training is the focus of \textit{New Benchmarks} and is additionally evaluated for \textit{New Responses} in Appendix \ref{app:new_responses}.

\textbf{New Tasks}: We hold out a subset of tasks and use IRT with task features (Section \ref{sec:irt-task}) to predict their difficulties, on every benchmark, reporting AUC-ROC averaged over 5-fold cross-validation. An ablation study in this setting isolates each feature source's contribution to difficulty prediction.

\textbf{New Responses}: We compare our IRT with agent features (Section \ref{sec:irt-agent}) against standard IRT on predicting held-out responses $y_{ij}$, without holding out any tasks or agents entirely, on every benchmark with 5-fold cross-validation.

\textbf{New Agents}: In each fold of a 5-fold cross-validation, we filter the held-out agents to those whose LLM and scaffold each appear in the training split, so both components are observed individually but not jointly. We use SWE-bench Verified and Terminal-Bench 2.0 because the other two benchmarks evaluate every LLM with the same fixed scaffold.

\textbf{New Benchmarks}: We train a multi-benchmark IRT model with task and agent features on three benchmarks and evaluate it on a fourth, held-out benchmark (filtering out agents with unseen LLMs or scaffolds). We use SWE-bench Pro and GSO as held-out benchmarks because SWE-bench Verified and Terminal-Bench 2.0 have many benchmark-specific LLMs and scaffolds.

\section{Results} 
\subsection{Agentic Task Artifacts Predict Task Difficulty Beyond the Problem Statement}
\label{sec:new-tasks} 

The \textit{New Tasks} setting reflects mid- to late-stage benchmark design: the designer has already evaluated agents on some tasks and wants to estimate difficulty for additional candidate tasks without running more evaluations. In Table \ref{tab:new-tasks}, we compare the held-out task performance using three different types of task feature vectors: LLM-as-a-judge features alone (with 15 features standardized across datasets), embedding features alone, and combined LLM-as-a-judge and embedding features. On all benchmarks, all three types of feature vectors significantly beat the baseline. Combining the feature vectors offers a small improvement over using individual vectors. For Terminal-Bench 2.0, the values in Table~\ref{tab:new-tasks} are based on collapsed (majority) responses; Table~\ref{tab:binomial-new-tasks} in Appendix~\ref{app:binomial} repeats this experiment with all attempts treated as independent trials and reaches the same qualitative conclusions.

\begin{table*}[h]
\small
\centering
\begin{tabular}{lcccc}
    \toprule
    Method & SWE-bench Verified & SWE-bench Pro & GSO & Terminal-Bench 2.0 \\
    \midrule
    Baseline & 0.718 $\pm$ 0.001 & 0.656 $\pm$ 0.001 & 0.718 $\pm$ 0.007 & 0.732 $\pm$ 0.002 \\
    Embedding & 0.824 $\pm$ 0.002 & 0.752 $\pm$ 0.003 & 0.756 $\pm$ 0.010 & 0.767 $\pm$ 0.010 \\
    LLM-as-a-judge & 0.840 $\pm$ 0.001 & 0.743 $\pm$ 0.003 & 0.778 $\pm$ 0.015 & \textbf{0.794} $\pm$ 0.007 \\
    Combined & \textbf{0.841} $\pm$ 0.003 & \textbf{0.763} $\pm$ 0.003 & \textbf{0.787} $\pm$ 0.014 & 0.792 $\pm$ 0.013 \\
    \\[-8pt]
    \hdashline
    \\[-7pt]
    Oracle & 0.945 $\pm$ 0.000 & 0.918 $\pm$ 0.000 & 0.916 $\pm$ 0.002 & 0.932 $\pm$ 0.001 \\
    \bottomrule
\end{tabular}
\caption{\textbf{AUC-ROC (mean $\pm$ std of 5-fold cross-validation means across 20 random seeds) on held-out tasks for each of the four benchmarks}. All feature vector-based predictors significantly beat the \textit{Baseline}, which always predicts the agent's empirical success rate from the training data. \textit{Oracle} is a standard IRT trained on all tasks, serving as an upper bound on performance.}
\label{tab:new-tasks}
\end{table*}

In a separate comparison in the same \textit{New Tasks} setting, we ablate the three feature sources specific to agentic tasks (repository state, test cases, and gold patch) to test whether each contributes information not provided by the others. We do not leave out the problem statement because the other three feature sources rely on it for context. Table~\ref{tab:information-ablation} reports the results: the full feature set improves substantially over problem-statement-only predictors on every benchmark, but the marginal value of any individual source is benchmark-dependent. On certain benchmarks, certain individual sources provide little additional information given the others. No source, however, is redundant on every benchmark: each contributes a meaningful gain on at least one dataset. We provide an additional (incremental) ablation setting in Appendix~\ref{app:hierarchical-ablation} and the prompts for the feature source ablation in Appendix \ref{app:info-ablation-prompts}.

\begin{table*}[h]
    \small
    \centering
    \setlength{\tabcolsep}{3pt}
    \begin{tabular}{@{}l@{\hskip 0pt}cccc@{}}
    \toprule
    Feature Source & SWE-bench Verified & SWE-bench Pro & GSO & Terminal-Bench 2.0 \\
    \midrule
    \textit{LLM-as-a-judge features} \\
    \quad Problem statement only & 0.786 $\pm$ 0.002 & 0.718 $\pm$ 0.002 & 0.710 $\pm$ 0.012 & 0.789 $\pm$ 0.007 \\
    \quad Full w/o repository state & 0.840 $\pm$ 0.001 & 0.732 $\pm$ 0.003 & \textbf{0.775} $\pm$ 0.017 & 0.797 $\pm$ 0.006 \\
    \quad Full w/o tests & 0.837 $\pm$ 0.001 & 0.735 $\pm$ 0.002 & 0.766 $\pm$ 0.011 & 0.798 $\pm$ 0.005 \\
    \quad Full w/o solution & 0.833 $\pm$ 0.002 & 0.748 $\pm$ 0.002 & 0.728 $\pm$ 0.010 & 0.797 $\pm$ 0.005 \\
    \quad Full & \textbf{0.845} $\pm$ 0.002 & \textbf{0.750} $\pm$ 0.003 & 0.774 $\pm$ 0.016 & \textbf{0.799} $\pm$ 0.006 \\
    \midrule
    \textit{Embedding features} \\
    \quad Problem Statement & 0.755 $\pm$ 0.004 & 0.726 $\pm$ 0.003 & 0.695 $\pm$ 0.050 & 0.776 $\pm$ 0.011 \\
    \quad \quad + Solution & \textbf{0.824} $\pm$ 0.002 & \textbf{0.752} $\pm$ 0.003 & \textbf{0.756} $\pm$ 0.010 & \textbf{0.767} $\pm$ 0.010 \\
    \bottomrule
    \end{tabular}
    \caption{\textbf{Feature source ablation: AUC-ROC (mean $\pm$ std of 5-fold cross-validation means across 20 random seeds) of predictors with some feature sources removed.} For LLM-as-a-judge features, we remove each of the three agentic feature sources one at a time. For embedding features, we compare a problem-statement-only predictor (the method of \citet{truong2025reliable} applied without modification) and one that additionally includes the solution.}
    \label{tab:information-ablation}
\end{table*}

\subsection{The LLM and Scaffold of an Agent Additively Predict Its Ability to Complete Tasks}
\label{sec:new-responses}

In the \textit{New Responses} setting, we validate our LLM-scaffold decomposition of agent ability by comparing IRT with agent features against standard IRT on held-out response prediction. We observe that the difference in performance between the two methods is small, and on Terminal-Bench 2.0, our method is even statistically indistinguishable from Standard IRT. This is a good result given that our method has higher generalization capabilities (it enables generalization to held-out agents, which we test in the \textit{New Agents} setting and which Standard IRT cannot do).

We provide the full \textit{New Responses} results in Appendix \ref{app:new_responses} and further validate the decomposition via a qualitative inspection of learned parameters (Appendix \ref{app:qualitative-inspection}) and a comparison of learned LLM abilities in multi-scaffold and fixed-scaffold settings (Appendix \ref{app:terminus}).

\subsection{Predictors with Task and Agent Features Generalize to Held-Out Agents and Benchmarks}

\label{sec:new-agents-benchmark}

\textbf{New Agents.} We observe very strong performance, even approaching the oracle, on held-out LLM-scaffold combinations (see Table \ref{tab:held-out-agent-results}). However, we remind the reader that both the LLM and the scaffold have individually been represented in the training data; our method cannot generalize to entirely new LLMs or scaffolds.

\begin{table}[h]
\small
\centering
\begin{tabular}{lcc}
\toprule
Method & SWE-bench Verified & Terminal-Bench 2.0 \\
\midrule
Baseline & 0.852 $\pm$ 0.010 & 0.839 $\pm$ 0.001 \\
IRT-Agent & \textbf{0.930} $\pm$ 0.002 & \textbf{0.923} $\pm$ 0.001 \\
\\[-8pt]
\hdashline
\\[-7pt]
Oracle & 0.949 $\pm$ 0.002 & 0.935 $\pm$ 0.001 \\
\bottomrule
\end{tabular}
\caption{\textbf{AUC-ROC (mean $\pm$ std of 5-fold cross-validation means across 20 random seeds) of predictors fitted on held-out agent data.} \textit{Baseline} predicts the task's empirical solve rate; \textit{Oracle} is a standard IRT trained on all data.}
\label{tab:held-out-agent-results}
\end{table}

\textbf{New Benchmarks.} This setting reflects the earliest stage of benchmark design, before any agent has been evaluated on the new benchmark. We observe decent generalization to held-out benchmarks, beating the baseline but still far from the oracle (see Table \ref{tab:multi-benchmark-results}). This gap is expected, as out-of-distribution generalization is much harder than the previous settings, and the predicted difficulties here should be read as a coarse first estimate, which a designer can refine as they collect partial agent evaluation data and switch to the \textit{New Tasks} setting.

\begin{table}[h]
\small
\centering
\begin{tabular}{lcc}
\toprule
Method & SWE-bench Pro & GSO \\
\midrule
Baseline & 0.571 & 0.637 \\
Embedding & 0.667 & 0.724 \\
LLM-as-a-judge & \textbf{0.696} & \textbf{0.743} \\
Combined & 0.674 & 0.719 \\
\\[-8pt]
\hdashline
\\[-7pt]
Oracle & 0.913 & 0.912 \\
\bottomrule
\end{tabular}
\caption{\textbf{AUC-ROC of predictors fitted on three benchmarks when evaluated on a fourth held-out benchmark.} \textit{Baseline} predicts the LLM's empirical success rate ignoring the scaffold; \textit{Oracle} is an IRT with agent features trained on all data from all four benchmarks.}
\label{tab:multi-benchmark-results}
\end{table}

\textbf{Application: Efficient Evaluation via Adaptive Task Selection.}
The \textit{New Agents} setting lets us estimate the ability of a new LLM-scaffold combination on an existing benchmark without running the full evaluation. We demonstrate this using computerized adaptive testing (CAT; \citealp{lord1968some}), which sequentially estimates an agent’s ability from the tasks evaluated so far and selects the next task to be maximally informative. CAT has been applied to non-agentic benchmarks by \citet{li2025adaptive}, \citet{truong2025reliable}, and \citet{hofmann2025fluid}.

We adapt CAT to agentic coding benchmarks by initializing the new agent’s ability estimate with our IRT-Agent model. Specifically, for a held-out LLM-scaffold pair $i$, we use the predicted ability $\hat{\theta}_i = \theta_{m} + \theta_{s}$ (with $\theta_m$ and $\theta_s$ derived from evaluations of other agents on the same benchmark) as the mean of a Gaussian prior. We then update the agent’s ability by MAP estimation after each observed task response. At each step, adaptive methods select the next task by maximizing Fisher information $I(\hat{\theta}_i; \beta_j) = \hat{p}_{ij}(1-\hat{p}_{ij})$, where $\hat{p}_{ij} = \sigma(\hat{\theta}_i - \beta_j)$.

We compare three strategies: (1) \textbf{Random + weak prior}, which samples tasks uniformly and uses a weakly informative $\mathcal{N}(0, 3^2)$ prior; (2) \textbf{Fisher + weak prior}, which adds Fisher-based task selection; and (3) \textbf{Fisher + IRT-Agent prior}, our method, which uses Fisher-based selection with the IRT-Agent prior, its variance set to the empirical RMSE of IRT-Agent ability predictions on held-out agents.

\begin{figure}[t]
    \centering
    \includegraphics[width=0.6\linewidth]{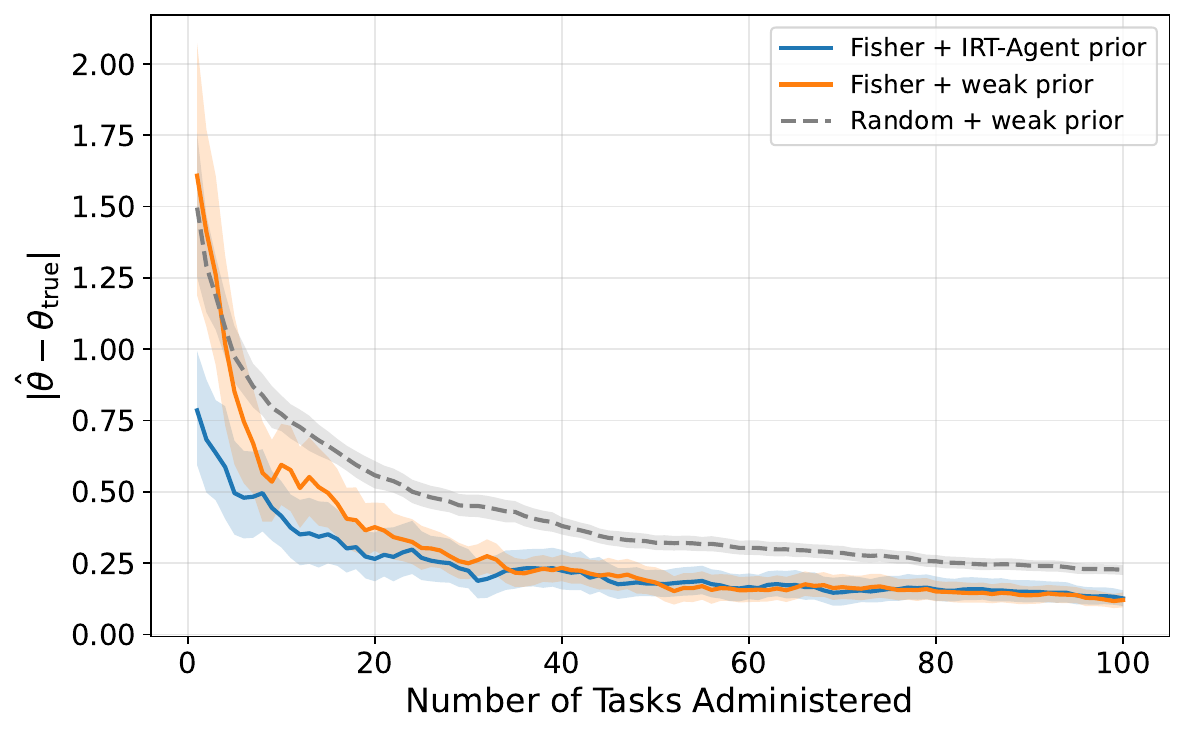}
    \caption{\textbf{Adaptive evaluation of a new agent on SWE-bench Verified.} Mean absolute error between the running ability estimate $\hat{\theta}_i$ and the full-benchmark IRT ability $\theta_i^\star$, averaged over 30 held-out LLM-scaffold pairs. Shaded regions show 95\% bootstrap confidence intervals over held-out agents.}
    \label{fig:adaptive_testing}
\end{figure}

We evaluate these strategies on SWE-bench Verified, holding out 30 LLM-scaffold pairs whose LLMs and scaffolds each appear elsewhere in the training data. For each held-out agent, we measure the absolute error between the running ability estimate $\hat{\theta}_i$ and the full-benchmark IRT ability $\theta_i^\star$. As shown in Figure~\ref{fig:adaptive_testing}, the IRT-Agent prior provides a strong initial estimate: before any tasks are administered, its mean absolute error is 0.79, which random selection only reaches after about 10 tasks. The advantage is largest at small task budgets; by roughly 40 tasks, the methods converge, as the observed responses eventually dominate the prior.

\section{Discussion}
\label{sec:discussion}

\textbf{Task Features.} For predicting the difficulty of a new task, agentic feature sources like repository state, test patches, and solution patches provide predictive power beyond the problem statement. This implies that difficulty should be measured differently for agentic coding tasks than for question-and-answer tasks, where the problem statement alone often conveys the full difficulty. However, our experiments are predictive, so we cannot tell whether the artifacts expose latent information already present in the problem statement or inherently generate difficulty. Our experimental design suggests the latter: our LLM-as-a-judge prompts are structured to capture properties of the agentic feature sources that the problem statement does not describe. Future work could probe causality by constructing counterfactual tasks that vary one aspect of these artifacts.

\textbf{Agent features.} A central contribution of our work is that agent ability decomposes additively into independent LLM and scaffold abilities, without pairwise interaction terms. This independence is what lets us compare agents across benchmarks and train a multi-benchmark IRT model. The decomposition also yields a novel quantitative ranking of scaffold abilities (Table \ref{tab:scaffold-abilities} in Appendix \ref{app:qualitative-inspection}). We caution that as more models are trained alongside specialized scaffolds, this independence may cease to hold. Our current method models LLMs and scaffolds as categorical features; we hope future work finds easily accessible quantitative or semantic agent features that generalize to fully unseen agents. Another direction is multi-agent systems: our framework excludes ensembles of multiple LLMs (Section \ref{sec:irt-agent}), but a natural extension is to assign a joint ability parameter to a specific ensemble and let inter-model interactions be absorbed into the scaffold ability term. 

\textbf{Applications of success prediction}. Predicting agent success at the task level has broad applications across the benchmark-design lifecycle, agent development, and training. For benchmark designers with evaluation data on some of their tasks (the \textit{New Tasks} setting), our predictors fill in difficulty estimates for the remaining tasks, supporting iteration on drafts and maintenance of evolving benchmarks. For designers with no evaluation data yet (the \textit{New Benchmarks} setting), they still provide a coarse first estimate of task difficulty. Agent developers can adaptively evaluate a new LLM-scaffold pair from a small subset of tasks, using its predicted ability from our IRT-Agent model as a prior in computerized adaptive testing (Section~\ref{sec:new-agents-benchmark}). Beyond evaluation, these predictors can guide task selection for reinforcement learning rollouts as in \citet{zheng2025act}: choosing tasks whose success probability is neither too low nor too high maximizes the advantage signal and improves training efficiency.

\textbf{Extension to richer IRT models.} Our framework supports binary task outcomes, including counts of repeated binary attempts via the binomial extension in Appendix \ref{app:binomial}, but not ordinal or continuous scores. IRT formulations such as the graded response model \citep{samejima1969estimation} and the partial credit model \citep{masters1982rasch} handle these settings, and extending our task-feature approach to them is a natural avenue for future work.

\section{Conclusion}

We introduce a method for predicting an agent’s probability of success on a coding task without requiring evaluation data for that task. We extend Item Response Theory, which represents agents and tasks with one-hot features and so cannot generalize to unseen agents or tasks: we replace categorical task identifiers with task features extracted by LLMs (embeddings or LLM-as-a-judge rubrics) from sources such as problem statements, repository states, test cases, and solution patches, and we encode each agent by its underlying LLM and scaffold. Our framework makes it cheaper for benchmark designers to estimate the difficulty of unevaluated tasks, and for agent developers to estimate the ability of new agents. It also quantifies scaffold effectiveness independently of the paired LLM, suggesting that the environment and grading setup play a substantial role in measured agent performance.

 \subsubsection*{Author Contributions}

\textbf{Chris Ge} and \textbf{Daria Kryvosheieva} led and executed most of the research and wrote the paper, jointly interpreting results and designing experiments to validate those interpretations. \textbf{Chris Ge} developed the LLM-as-a-judge features, implemented the \textit{New Tasks}, feature source ablations, and adaptive testing experiments, conducted the literature review, and formulated the narrative framing of the paper. \textbf{Daria Kryvosheieva}  developed the embedding features,  proposed and investigated the LLM-scaffold decomposition for agent abilities,  and implemented the \textit{New Responses}, \textit{New Agents}, and \textit{New Benchmarks}  experiments. \textbf{Kaivalya Hariharan} acted as the primary supervisor for the technical and framing aspects of the project, helped create the initial proposal, implemented the IRT model training, and edited the paper. \textbf{Uzay Girit} helped with the initial proposal, assisted in interpreting intermediate results, and provided feedback on the paper. \textbf{Daniel Fried} provided feedback on the initial proposal as well as the paper.

\subsubsection*{Acknowledgments}

We thank \textbf{Zifan Carl Guo} for his valuable discussion throughout the project and feedback on the paper. We thank \textbf{Tony Wang} for providing feedback on the paper as well.

\bibliography{colm2026_conference}
\bibliographystyle{colm2026_conference}

\appendix

\section{Details on Task Feature Extraction}
\label{app:details-task-feature-extraction}


\subsection{Embedding Features}
\label{app:embedding-features}

We embed the task, followed by its ``gold'' solution and an instruction sentence, using an instruction-tuned LLM. We extract embeddings from the last layer's hidden state via last-token pooling, as this is the standard practice with autoregressive LLMs \citep{zhang-etal-2025-language, zhang2025qwen3embeddingadvancingtext, li2025making}. We use \texttt{DeepSeek-R1-Distill-Qwen-32B} \citep{Guo2025} as the embedding backbone, as we found it to produce the best embeddings among the 17 backbones we tested (see Appendix \ref{app:backbone}). We provide the input template in Appendix \ref{app:embedding-prompts} and verify the usefulness of the solution through an ablation (see Table \ref{tab:information-ablation}).

\subsection{LLM-as-a-Judge Features}
\label{app:llm-judge-features}

We prompt Claude Opus 4.6 \citep{anthropic-2026} to assess tasks along a fixed rubric of manually specified criteria, such as the complexity of the required fix or the ease of verifying its correctness. For each criterion, we use a discrete scale. We chose Claude Opus 4.6 according to a judge model ablation experiment detailed in Appendix \ref{app:llm_judge_backbone}.

We use two separate prompt templates for (i) most feature sources (problem statement, test cases, solution) and (ii) repository state features. Both templates are provided in Appendix \ref{app:llm-judge-prompts}. The former (main) template includes all three of the problem statement, test cases, and solution, regardless which of the these feature sources are being extracted. Metadata about the task (like its category for Terminal-Bench 2.0 or which repository it comes from for SWE-Bench Verified) are included in the prompt as well. To extract the repository state features, we run GPT-5.4 (used instead of Claude Opus 4.6 for cost reasons) as an ``auditor agent'' on the task itself through the InspectAI framework \citep{ukaisi2024inspect}, overriding the task instructions in order to have the agent explore the task sandbox, using a ReAct-style coding scaffold of a bash shell and a Python interpreter with a limit of $100$ messages and a $240$-second per-tool-call timeout.

In total, we extract $28$ features: $15$ about the problem statement, $3$ about the test cases, $2$ about the solution, and $8$ about the repository state. However, we do not use all 28 features in every experiment, because some of them are less useful or only useful for specific benchmarks. Instead, every experiment uses some subset of 15 of these features: for all experiments except the feature source ablations, we use the same fixed 15-feature subset, while for the ablations, we select the top-15 features by ridge regression coefficient at each ablation level. The default (non-ablation) feature set contains 10 features about the problem statement, 1 about the test cases, 1 about the solution, and 3 about repository state; these features were chosen manually to be consistently useful across benchmarks and are listed in Appendix \ref{app:llm-judge-prompts}. To mitigate overfitting bias, we made this selection using a different random seed from the one that we ended up using for the \textit{New Tasks} experiment. This default set is identical for every benchmark: the only per-dataset variation is in the wording of some feature scales (Table \ref{tab:scale-variants}), and we treat such variants as the same feature. Table \ref{tab:judge-feature-inventory} summarizes which features were extracted and which ones were used in the default set.

\begin{table}[h]
\small
\centering
\begin{tabular}{llcc}
\toprule
Feature source & Extracted by & \# extracted & \# in default set \\
\midrule
Problem statement & Claude Opus 4.6 (main prompt) & 15 & 10 \\
Test cases & Claude Opus 4.6 (main prompt) & 3 & 1 \\
Solution & Claude Opus 4.6 (main prompt) & 2 & 1 \\
Repository state & GPT-5.4 (auditor agent prompt) & 8 & 3 \\
\midrule
Total & & 28 & 15 \\
\bottomrule
\end{tabular}
\caption{\textbf{LLM-as-a-judge feature inventory.} The default set of 15 features is used, unchanged, for every benchmark and every experiment except the feature source ablations.}
\label{tab:judge-feature-inventory}
\end{table}

The feature source ablations are the only experiments that deviate from the default set. Normally, the prompts for the extraction of problem statement, test case, and solution features include information about all three of these feature sources, but to ensure a clean ablation, we completely remove a feature source when it is being ablated---that includes re-extracting features for other sources with the ablated source(s) excluded from the prompt. (The repository state features are not re-extracted because their extraction prompt does not use information from other feature sources.) For each ablation level, we then select the top $15$ features with the largest (by magnitude) ridge regression coefficient from all features of the included sources. Table \ref{tab:ablation-feature-versions} lists the configuration of every ablation level.


\begin{table}[h]
\small
\centering
\setlength{\tabcolsep}{5pt}
\begin{tabular}{llll}
\toprule
Ablation  (Table \ref{tab:information-ablation}) & Ablation  (Table \ref{tab:hierarchical-ablation}) & Sources in predictor & Sources in main prompt \\
\midrule
Problem statement only & Problem statement & stmt. & stmt. \\
 & + Repository state & stmt., repo. & stmt. \\
Full w/o solution & + Tests & stmt., tests, repo. & stmt., tests \\
Full w/o tests & & stmt., sol., repo. & stmt., sol. \\
Full w/o repository state & & stmt., tests, sol. & stmt., tests, sol. \\
Full & + Solution & stmt., tests, sol., repo. & stmt., tests, sol. \\
\bottomrule
\end{tabular}
\caption{\textbf{Configuration of each feature source ablation level.} ``Sources in predictor'' lists the sources whose features enter the predictor at the given level. ``Sources in main prompt'' lists the sources provided in the main judge prompt (used to extract all features included at the given level except the repository state features, which are extracted via an independent pipeline).}
\label{tab:ablation-feature-versions}
\end{table}
\section{An Alternative Interpretation of Agent Features}
\label{app:agent_features_one_hot}

In sections \ref{sec:irt-task} and \ref{sec:irt-agent}, we seemingly incorporate task features (numerical vectors extracted using embeddings or LLM-as-a-judge) and agent features (the categorical LLM and scaffold) differently  into the IRT model. For the former, we replaced the IRT difficulty with a linear combination of the task features. For the latter,  we directly replaced the IRT agent abilities by new ability parameters, one for each LLM and one for each scaffold. We now point out  that introducing new LLM and scaffold abilities is mathematically equivalent to using a linear combination of the concatenated one-hot vectors for LLMs and scaffolds. If LLM $m$ has ability $\theta_m$ and scaffold $s$ has ability $\theta_s$, then setting the weight corresponding to the index for $m$ on the concatenated one-hot vector to $\theta_m$ and the weight corresponding to the index for $s$ to $\theta_s$ results in the same estimated agent ability $\theta_m + \theta_s$. This mapping is very easily reversible, so it is a bijection. 

This perspective restores the symmetry in our treatment of task features and agent features in the functional form  of IRT.  The only remaining difference is the way we choose to train the parameters: we train using a ridge regression for task features, while we train directly to optimize log-likelihood for agent features. This choice was made empirically, and we did try training to optimize log likelihood for task features in Appendix \ref{app:joint_difficulty_training}.

\section{Ablations}

\subsection{Embedding Backbone}
\label{app:backbone}

We tested 17 open-weight instruction-tuned LLMs as embedding backbones on held-out tasks in SWE-bench Verified (see Table \ref{tab:backbone}), and selected the backbone with the best performance (\texttt{DeepSeek-R1-Distill-Qwen-32B}).

\begin{table}[h]
\small
\centering
\small
\begin{tabular}{lc}
\toprule
Backbone & AUC-ROC $\pm$ std \\
\midrule
\texttt{Qwen3-VL-4B-Instruct} & 0.814 $\pm$ 0.017 \\
\texttt{Qwen3-VL-8B-Instruct} & 0.816 $\pm$ 0.017 \\
\texttt{Qwen3-VL-32B-Instruct} & 0.817 $\pm$ 0.018 \\
\texttt{Qwen3-8B} & 0.798 $\pm$ 0.021 \\
\texttt{Qwen3-14B} & 0.817 $\pm$ 0.021 \\
\texttt{Qwen3-32B} & 0.801 $\pm$ 0.026 \\
\texttt{Qwen2.5-Coder-7B-Instruct} & 0.803 $\pm$ 0.023 \\
\texttt{Qwen2.5-Coder-14B-Instruct} & 0.807 $\pm$ 0.016 \\
\texttt{Qwen2.5-Coder-32B-Instruct} & 0.809 $\pm$ 0.022 \\
\texttt{gemma-3-4b-it} & 0.792 $\pm$ 0.014 \\
\texttt{gemma-3-12b-it} & 0.812 $\pm$ 0.021 \\
\texttt{gemma-3-27b-it} & 0.815 $\pm$ 0.019 \\
\texttt{DeepSeek-R1-Distill-Qwen-7B} & 0.801 $\pm$ 0.018 \\
\texttt{DeepSeek-R1-Distill-Llama-8B} & 0.807 $\pm$ 0.025 \\
\texttt{DeepSeek-R1-Distill-Qwen-14B} & 0.821 $\pm$ 0.020 \\
\texttt{DeepSeek-R1-Distill-Qwen-32B} & \textbf{0.824} $\pm$ 0.020 \\
\texttt{Llama-3.2-11B-Vision-Instruct} & 0.791 $\pm$ 0.022 \\
\bottomrule
\end{tabular}
\caption{\textbf{Embedding backbone test}. \textit{AUC-ROC} is averaged over 5 cross-validation folds; standard deviation (\textit{std}) is also shown.}
\label{tab:backbone}
\end{table}

\subsection{LLM-as-a-Judge Model}
\label{app:llm_judge_backbone}

We ablate the model used to extract the 12 non-repository-state LLM-as-a-judge features, keeping the same choice of 15 features. The 3 repository state features, extracted by GPT-5.4 via the auditor agent, are kept constant. Table~\ref{tab:llm_judge_backbone} shows results for Claude Opus 4.6 (the default), GPT-5.4, and Claude Sonnet 4.6. We chose Claude Opus 4.6 as it has reasonable performance across all datasets.  

\begin{table}[h]
\small
\centering
\begin{tabular}{lccc}
\toprule
Benchmark & Claude Opus 4.6 & GPT-5.4 & Claude Sonnet 4.6 \\
\midrule
SWE-bench Verified & \textbf{0.842} & 0.835 & 0.838 \\
SWE-bench Pro & 0.759 & \textbf{0.760} & 0.758 \\
GSO & \textbf{0.804} & 0.701 & 0.746 \\
Terminal-Bench 2.0 & 0.810 & 0.828 & \textbf{0.830} \\
\bottomrule
\end{tabular}
\caption{\textbf{LLM-as-a-Judge backbone ablation.} Grouped Ridge (Embedding + LLM Judge) AUC-ROC across backbones. Bold indicates best per dataset.}
\label{tab:llm_judge_backbone}
\end{table}

\subsection{Hierarchical Feature Source Ablation}
\label{app:hierarchical-ablation}

The four feature sources have a natural ordering by how widely each artifact is available. The \textit{problem statement} is universal: every benchmark and every agent has it. \textit{Repository state} can be obtained at runtime by any agent willing to explore the task sandbox. \textit{Test cases} are held by every benchmark designer because they are required to grade submissions, but are not always exposed to the agent. Reference \textit{solutions} are an additional artifact that only some benchmarks publish. Table~\ref{tab:hierarchical-ablation} traces predictor performance as each source is added; it complements the leave-one-out view in Table~\ref{tab:information-ablation}.

\begin{table}[h]
\small
\centering
\begin{tabular}{lcccc}
\toprule
Feature Source & SWE-bench Verified & SWE-bench Pro & GSO & Terminal-Bench 2.0 \\
\midrule
Problem statement & 0.786 $\pm$ 0.002 & 0.718 $\pm$ 0.002 & 0.710 $\pm$ 0.012 & 0.789 $\pm$ 0.007 \\
+ Repository state & 0.797 $\pm$ 0.002 & 0.735 $\pm$ 0.001 & 0.713 $\pm$ 0.012 & 0.796 $\pm$ 0.008 \\
+ Tests & 0.833 $\pm$ 0.002 & 0.748 $\pm$ 0.002 & 0.728 $\pm$ 0.010 & 0.797 $\pm$ 0.005 \\
+ Solution & \textbf{0.845} $\pm$ 0.002 & \textbf{0.750} $\pm$ 0.003 & \textbf{0.774} $\pm$ 0.016 & \textbf{0.799} $\pm$ 0.006 \\
\bottomrule
\end{tabular}
\caption{\textbf{Hierarchical information source ablation.} AUC-ROC (mean $\pm$ std of cross-validation means across 20 random seeds) of LLM-as-a-judge difficulty predictors as feature sources are added in order of increasing access (problem statement, repository state, test cases, solution). Each row adds one source on top of those in the previous row, and the top 15 features are selected by ridge regression coefficient from the cumulative features  available at each row, similar to  Table~\ref{tab:information-ablation}. The solution contributes the largest gain on three of four benchmarks---most strikingly on GSO, where adding the  solution lifts AUC from 0.728 to 0.774---but it is also the feature source which is the most difficult to obtain. On SWE-bench Verified, the test cases contribute a comparable amount (0.797~$\to$~0.833).}
\label{tab:hierarchical-ablation}
\end{table}

Performance is non-decreasing at each step on every benchmark. The takeaway is straightforward: a difficulty predictor should be given as much information as is available, since adding any source never hurts and the most privileged sources tend to help the most. This complements the leave-one-out view in Table~\ref{tab:information-ablation}, which shows that  no source is redundant on every benchmark.

\subsection{Joint Log-Likelihood Maximization}
\label{app:joint_difficulty_training}

Below, we show the results of the same experimental setup as Table \ref{tab:new-tasks} but when we train the linear model difficulty predictor weights jointly with the IRT abilities instead of freezing learned IRT ability and difficulty scores and training a ridge regression. We find that this training results in slightly worse AUC-ROCs, so in our main experiments, we use ridge regression. 

\begin{table}[h]
    \centering
    \begin{tabular}{lcccc:c}
    \toprule
    Benchmark & Baseline & Embedding & LLM-as-a-Judge & Combined & Oracle \\
    \midrule
    SWE-bench Verified & 0.718 & 0.825 & 0.842 & \textbf{0.843} & 0.945 \\
    SWE-bench Pro & 0.657 & \textbf{0.756} & 0.744 & 0.750 & 0.918 \\
    GSO & 0.714 & 0.771 & 0.769 & \textbf{0.781} & 0.914 \\
    Terminal-Bench 2.0 & 0.734 & 0.795 & 0.805 & \textbf{0.807} & 0.932 \\
    \bottomrule
    \end{tabular}
        \caption{\textbf{Jointly training the linear difficulty predictor model with the IRT abilities} results in slightly lower AUC  for held-out tasks on most benchmarks  compared to training a ridge regression and using frozen IRT abilities during evaluation. }
    \label{tab:joint_difficulty_predictor_training}
\end{table}

\subsection{Functional Forms Relating the LLM and Scaffold Abilities}
\label{app:functional-forms}

We tried multiple ways to derive the agent's ability from its LLM and scaffold abilities. We tested each functional form on held-out responses in SWE-bench Verified (Table \ref{tab:functional-forms}) and found summation to perform best on cross-validation AUC-ROC.

\begin{table}[h]
\small
\centering
\small
\begin{tabular}{lcc}
\toprule
Functional Form & Formula & AUC-ROC \\
\midrule
Sum & $\theta_m + \theta_s$ & \textbf{0.939} $\pm$ 0.000 \\
Maximum & $\max(\theta_m, \theta_s)$ & 0.937 $\pm$ 0.000 \\
Minimum & $\min(\theta_m, \theta_s)$ & 0.937 $\pm$ 0.000 \\
Product & $\mathrm{sign}(\theta_m + \theta_s) \left|\theta_m \theta_s\right|$ & 0.933 $\pm$ 0.001 \\
L2 norm & $\mathrm{sign}(\theta_m + \theta_s)\sqrt{\theta_m^2 + \theta_s^2}$ & 0.938 $\pm$ 0.000 \\
Harmonic mean & $\mathrm{sign}(\theta_m + \theta_s) \left|\frac{2 \theta_m \theta_s}{\theta_m + \theta_s}\right|$ & 0.815 $\pm$ 0.004 \\
Weighted sum & $\frac{w_m}{w_m + w_s}\theta_m + \frac{w_s}{w_m+w_s}\theta_s$ & \textbf{0.939} $\pm$ 0.000 \\
Sum with variance term & $\theta_m + \lambda_m \theta_s$ & \textbf{0.939} $\pm$ 0.000 \\
\bottomrule
\end{tabular}
\caption{\textbf{Functional form test for the relationship between the LLM and scaffold ability parameters}. All runs are conducted with the best backbone (\texttt{DeepSeek-R1-Distill-Qwen-32B}). \textit{AUC-ROC} is the mean ($\pm$ std) of 5-fold cross-validation means across 20 random seeds. In the weighted sum, $w_m$ and $w_s$ are additional learnable parameters (shared for all LLMs and scaffolds). Sum with variance term aims to capture each LLM's sensitivity to scaffolds: combined with a good scaffold (positive $\theta_s$), an LLM with a high variance term $\lambda_m$ will do better than predicted by regular sum, and combined with a bad scaffold (negative $\theta_s$), worse than predicted by regular sum.}
\label{tab:functional-forms}
\end{table}

\subsection{All Attempts as Independent Trials for Terminal-Bench 2.0}
\label{app:binomial}

In the main paper (Section~\ref{sec:agentic-coding-benchmarks}), we take the majority outcome of each agent's multiple attempts on a Terminal-Bench 2.0 task as a single binary response. Here we instead use each attempt as its own observation, both when training the IRT and when evaluating its predictions. The IRT model that predicts the probability of a single successful attempt (Equation~\ref{eq:irt-prediction}) is unchanged; only the data-fit term in the SVI training objective and the data used during evaluation change.

\textbf{Training change.} Standard IRT training uses the Bernoulli log likelihood as the data-fit term in the SVI objective,
\begin{equation}
\label{eq:bernoulli-loglik}
\mathcal{L}_{\text{Bern}}(\theta, \beta) = \sum_{i, j} y_{ij} \log \sigma(\theta_i - \beta_j) + (1 - y_{ij}) \log\!\bigl(1 - \sigma(\theta_i - \beta_j)\bigr),
\end{equation}
where $y_{ij} \in \{0, 1\}$ is agent $i$'s majority outcome on task $j$. We replace this with a Binomial data-fit term that uses each attempt individually: if agent $i$ succeeds on $k_{ij}$ of $n_{ij}$ attempts on task $j$,
\begin{equation}
\label{eq:binomial-loglik}
\mathcal{L}_{\text{Bin}}(\theta, \beta) = \sum_{i, j} k_{ij} \log \sigma(\theta_i - \beta_j) + (n_{ij} - k_{ij}) \log\!\bigl(1 - \sigma(\theta_i - \beta_j)\bigr).
\end{equation}
This formulation assumes that, for a given agent and task, the $n_{ij}$ attempts are independent and each succeeds with the same probability: the IRT-predicted probability $\sigma(\theta_i - \beta_j)$ from Equation~\ref{eq:irt-prediction}. The ridge regression that maps task features to difficulties (Equation~\ref{eq:ridge}) is unchanged, and so is the inference rule (Equation~\ref{eq:irt-task-prediction}).

\textbf{Evaluation change.} We compute AUC-ROC over all attempts rather than over majority outcomes. This means that for the all attempts case,  for any agent-task pair with a mix of successful and failed attempts, the IRT predicts a single probability that has to be ranked against both positive and negative observations from that pair.
Empirically (Table~\ref{tab:binomial-new-tasks}), only the Oracle's AUC drops meaningfully under this protocol; the predictor methods reach comparable AUC to the binary setting.

\textbf{Results.} Table~\ref{tab:binomial-new-tasks} shows the Terminal-Bench 2.0 column of Table~\ref{tab:new-tasks} under this protocol.

\begin{table}[h]
    \small
    \centering
    \setlength{\tabcolsep}{3pt}
    \begin{tabular}{l@{\hskip 2pt}cccc:c}
    \toprule
    Protocol & Baseline & Embedding & LLM-as-a-Judge & Combined & Oracle \\
    \midrule
    Binary (main paper) & 0.732 $\pm$ 0.002 & 0.767 $\pm$ 0.010 & \textbf{0.794} $\pm$ 0.007 & 0.792 $\pm$ 0.013 & 0.932 $\pm$ 0.001 \\
    All attempts & 0.754 $\pm$ 0.002 & 0.777 $\pm$ 0.005 & \textbf{0.801} $\pm$ 0.005 & 0.796 $\pm$ 0.010 & 0.912 $\pm$ 0.001 \\
    \bottomrule
    \end{tabular}
    \caption{\textbf{Treating each attempt as an independent trial on Terminal-Bench 2.0} (mean $\pm$ std of cross-validation means across 20 random seeds). The \emph{Binary} row reproduces the Terminal-Bench 2.0 row of Table~\ref{tab:new-tasks}; the \emph{All attempts} row repeats the same experiment but treats each attempt as an independent observation for both training and evaluation. The Oracle drops by about 0.02 AUC under this protocol due to being evaluated on multiple attempts instead of one  binary majority outcome; the predictor methods reach comparable AUC under both protocols.}
    \label{tab:binomial-new-tasks}
\end{table}

As in the main paper, all of our methods significantly beat the baseline.

\section{AUC-ROC Metric Description}
\label{app:AUC}

The area under the receiver operating characteristic curve, or AUC-ROC, is a number from $0$ to $1$ that measures how well the model captures true positives while avoiding false positives, with 1 being a perfectly accurate model and 0.5 being the expected score of random guessing. The other way to interpret AUC-ROC is: given a true successful agent run and a true failing run, what is the probability that the success probability predicted for the successful run is higher than the success probability predicted for the failing run. 

The AUC-ROC has two advantages over accuracy: it is robust to class imbalance, and it doesn't require a fixed threshold on the predicted probability between predicting success and predicting failure.

\section{Bootstrapping}
\label{app:bootstrap}

In experiments where we use 5-fold cross-validation (\textit{New Tasks}, \textit{New Responses}, and \textit{New Agents}), we consider $N=20$ random seeds determining fold splits. For each of the $N$ seeds, for each IRT-based method, we compute the difference between the mean AUCs (taken across 5 folds) of the method and the baseline: $\Delta_{\text{seed}, \text{method}} = \text{AUC}(\text{method}) - \text{AUC}(\text{baseline})$. For each method, we then take $B=10,000$ bootstrapped samples (with replacement) of size $N$ of the $N$ differences $\Delta_{\text{seed}, \text{method}}$, and compute the 95\% confidence interval for the mean of the bootstrapped sample. In the \textit{New Benchmarks} experiment, which does not use cross-validation, we instead perform bootstrapping clustered by task.

Each experiment reports several benchmark and method comparisons against the baseline, so we also apply a Holm-Bonferroni correction to control the family-wise error rate within each experiment. The family is the full set of comparisons in the experiment's table: 12 for \textit{New Tasks}, 4 for \textit{New Responses}, 2 for \textit{New Agents}, and 6 for \textit{New Benchmarks}. The tables below report both the raw and the adjusted $p$-values. Every comparison that is significant at the 0.05 level before the correction remains significant after it.

\subsection{New Tasks}
\label{app:bootstrap-new-tasks}

\begin{table}[h]
    \small
    \centering
    
    \begin{subtable}[b]{\textwidth}
        \centering
        \begin{tabular}{lcccc}
            \toprule
            Method & Delta & 95\% CI & $p$ & Adjusted $p$ \\
            \midrule
            Embedding & 0.106 & [0.105, 0.107] & $<0.0002$ & $<0.0024$ \\
            LLM-as-a-judge & 0.123 & [0.122, 0.123] & $<0.0002$ & $<0.0024$ \\
            Combined & 0.123 & [0.122, 0.124] & $<0.0002$ & $<0.0024$ \\
        \end{tabular}
        \caption{SWE-bench Verified}
        \label{tab:bootstrap-new-tasks-verified}
    \end{subtable}
    
    \begin{subtable}[b]{\textwidth}
        \centering
        \begin{tabular}{lcccc}
            \toprule
            Method & Delta & 95\% CI & $p$ & Adjusted $p$ \\
            \midrule
            Embedding & 0.095 & [0.094, 0.097] & $<0.0002$ & $<0.0024$ \\
            LLM-as-a-judge & 0.087 & [0.085, 0.088] & $<0.0002$ & $<0.0024$ \\
            Combined & 0.106 & [0.105, 0.108] & $<0.0002$ & $<0.0024$ \\
        \end{tabular}
        \caption{SWE-bench Pro}
        \label{tab:bootstrap-new-tasks-pro}
    \end{subtable}
    
    \begin{subtable}[b]{\textwidth}
        \centering
        \begin{tabular}{lcccc}
            \toprule
            Method & Delta & 95\% CI & $p$ & Adjusted $p$ \\
            \midrule
            Embedding & 0.039 & [0.035, 0.042] & $<0.0002$ & $<0.0024$ \\
            LLM-as-a-judge & 0.061 & [0.054, 0.067] & $<0.0002$ & $<0.0024$ \\
            Combined & 0.069 & [0.064, 0.075] & $<0.0002$ & $<0.0024$ \\
        \end{tabular}
        \caption{GSO}
        \label{tab:bootstrap-new-tasks-gso}
    \end{subtable}
    
    \begin{subtable}[b]{\textwidth}
        \centering
        \begin{tabular}{lcccc}
            \toprule
            Method & Delta & 95\% CI & $p$ & Adjusted $p$ \\
            \midrule
            Embedding & 0.034 & [0.030, 0.039] & $<0.0002$ & $<0.0024$ \\
            LLM-as-a-judge & 0.062 & [0.059, 0.062] & $<0.0002$ & $<0.0024$ \\
            Combined & 0.059 & [0.054, 0.065] & $<0.0002$ & $<0.0024$ \\
        \end{tabular}
        \caption{Terminal-Bench 2.0}
        \label{tab:bootstrap-new-tasks-terminalbench}
    \end{subtable}
    
    \caption{\textbf{Bootstrapping results for the \textit{New Tasks} experiment.} All IRT-based methods significantly beat the naive baseline (two-sided $p$-value below the resolution of bootstrapping with 10,000 samples, which is equal to 0.0002; Holm-adjusted $p$-value below 0.0024).}
    \label{tab:bootstrap-new-tasks}
\end{table}

\subsection{New Responses}
\label{app:bootstrap-new-reponses}

\begin{table}[H]
    \small
    \centering
    
    \begin{subtable}[b]{\textwidth}
        \centering
        \begin{tabular}{lcccc}
            \toprule
            Method & Delta & 95\% CI & $p$ & Adjusted $p$ \\
            \midrule
            IRT-Agent & -0.001 & [-0.001, -0.001] & $<0.0002$ & $<0.0008$ \\
        \end{tabular}
        \caption{SWE-bench Verified}
        \label{tab:bootstrap-new-responses-verified}
    \end{subtable}
    
    \begin{subtable}[b]{\textwidth}
        \centering
        \begin{tabular}{lcccc}
            \toprule
            Method & Delta & 95\% CI & $p$ & Adjusted $p$ \\
            \midrule
            IRT-Agent & -0.005 & [-0.005, -0.005] & $<0.0002$ & $<0.0008$ \\
        \end{tabular}
        \caption{SWE-bench Pro}
        \label{tab:bootstrap-new-responses-pro}
    \end{subtable}
    
    \begin{subtable}[b]{\textwidth}
        \centering
        \begin{tabular}{lcccc}
            \toprule
            Method & Delta & 95\% CI & $p$ & Adjusted $p$ \\
            \midrule
            IRT-Agent & -0.004 & [-0.006, -0.003] & $<0.0002$ & $<0.0008$ \\
        \end{tabular}
        \caption{GSO}
        \label{tab:bootstrap-new-responses-gso}
    \end{subtable}
    
    \begin{subtable}[b]{\textwidth}
        \centering
        \begin{tabular}{lcccc}
            \toprule
            Method & Delta & 95\% CI & $p$ & Adjusted $p$ \\
            \midrule
            IRT-Agent & 0.000 & [0.000, 0.000] & 0.1462 & 0.1462 \\
        \end{tabular}
        \caption{Terminal-Bench 2.0}
        \label{tab:bootstrap-new-responses-terminalbench}
    \end{subtable}
    
    \caption{\textbf{Bootstrapping results for the \textit{New Responses} experiment.} \textit{IRT-Agent}---the proposed IRT model with LLM-scaffold decomposition---is not expected to (significantly) beat \textit{Standard IRT}, because its intended use case is generalization to multi-benchmark IRT models.}
    \label{tab:bootstrap-new-responses}
\end{table}

\subsection{New Agents}
\label{app:bootstrap-new-agents}

\begin{table}[h]
    \small
    \centering
    
    \begin{subtable}[b]{\textwidth}
        \centering
        \begin{tabular}{lcccc}
            \toprule
            Method & Delta & 95\% CI & $p$ & Adjusted $p$ \\
            \midrule
            IRT-Agent & 0.079 & [0.074, 0.084] & $<0.0002$ & $<0.0004$ \\
        \end{tabular}
        \caption{SWE-bench Verified}
        \label{tab:bootstrap-new-agents-verified}
    \end{subtable}
    
    \begin{subtable}[b]{\textwidth}
        \centering
        \begin{tabular}{lcccc}
            \toprule
            Method & Delta & 95\% CI & $p$ & Adjusted $p$ \\
            \midrule
            IRT-Agent & 0.084 & [0.083, 0.085] & $<0.0002$ & $<0.0004$ \\
        \end{tabular}
        \caption{Terminal-Bench 2.0}
        \label{tab:bootstrap-new-agents-terminalbench}
    \end{subtable}
    
    \caption{\textbf{Bootstrapping results for the \textit{New Agents} experiment.} \textit{IRT-Agent} significantly beats the naive baseline (two-sided $p$-value below the resolution of bootstrapping with 10,000 samples, which is equal to 0.0002; Holm-adjusted $p$-value below 0.0004).}
    \label{tab:bootstrap-new-agents}
\end{table}

\subsection{New Benchmarks}
\label{app:bootstrap-new-benchmarks}

\begin{table}[H]
    \small
    \centering
    
    \begin{subtable}[b]{\textwidth}
        \centering
        \begin{tabular}{lcccc}
            \toprule
            Method & Delta & 95\% CI & $p$ & Adjusted $p$ \\
            \midrule
            Embedding & 0.096 & [0.074, 0.117] & $<0.0002$ & $<0.0012$ \\
            LLM-as-a-judge & 0.125 & [0.101, 0.147] & $<0.0002$ & $<0.0012$ \\
            Combined & 0.102 & [0.079, 0.125] & $<0.0002$ & $<0.0012$ \\
        \end{tabular}
        \caption{Held-out SWE-bench Pro}
        \label{tab:bootstrap-new-benchmarks-pro}
    \end{subtable}
    
    \begin{subtable}[b]{\textwidth}
        \centering
        \begin{tabular}{lcccc}
            \toprule
            Method & Delta & 95\% CI & $p$ & Adjusted $p$ \\
            \midrule
            Embedding & 0.087 & [0.028, 0.141] & 0.0070 & 0.0140 \\
            LLM-as-a-judge & 0.105 & [0.040, 0.166] & 0.0026 & 0.0078 \\
            Combined & 0.082 & [0.010, 0.149] & 0.0280 & 0.0280 \\
        \end{tabular}
        \caption{Held-out GSO}
        \label{tab:bootstrap-new-benchmarks-gso}
    \end{subtable}
    
    \caption{\textbf{Bootstrapping results for the \textit{New Benchmarks} experiment.} All IRT-based methods significantly beat the naive baseline, both before and after the Holm-Bonferroni correction.}
    \label{tab:bootstrap-new-benchmarks}
\end{table}

\section{Prompts}
\label{app:prompts}

\subsection{Embedding Prompts}
\label{app:embedding-prompts}


Below is the prompt template we use to extract embeddings for all tasks.

\begin{promptbox}[title=Embedding Prompt Template]
INSTRUCTION = "How difficult is the above task for a coding agent? Please output one floating-point number from 0 (very easy) to 1 (very hard). Your difficulty score:\n"

def prompt(statement, solution, instruction=INSTRUCTION):
    return f"Task statement:\n{statement}\n\nSolution:{solution}\n\n{instruction}"
\end{promptbox}

We use a sequence length of 8192 with left truncation (i.e., prompts exceeding the length limit are truncated to the last 8192 tokens).

\subsection{LLM-as-a-Judge Prompts}
\label{app:llm-judge-prompts}

Each prompt is composed of a dataset-specific \textbf{introduction}, the \textbf{task information} including all feature sources used (presented here as an f-string template), the definitions of the \textbf{scale} of the discrete features, and an \textbf{output format} specification. Some feature scales are slightly modified between datasets to more accurately reflect what that feature means in the context of the dataset; most commonly, GSO has different feature scale definitions because it is about optimization, not bug fixing. However, we emphasize that these features still represent the same high level idea, and we treat them as the same feature across different datasets in the multi-benchmark experiments. Features are extracted in batches of at most $7$ per API call, using prefix caching between consecutive batches of features. 

\subsubsection{Main Feature Extraction Prompt}
\label{app:main-feature-prompt}

Below is the prompt structure for the $12$ non-environment features used in all experiments (10 statement features, 1 test feature, and 1 solution feature). All features are extracted using Claude Opus 4.6 with the problem statement, test cases, and solution all included in the prompt. We show the SWE-bench Verified (``code'') variant for defining the scales of the features; per-dataset scale variants for Terminal-Bench 2.0 and GSO are provided in Table \ref{tab:scale-variants}.

The \textbf{introduction} varies by dataset. We show all four variants with all feature sources available in the prompt:

\begin{promptbox}[title=Introduction --- SWE-bench Verified]
You are analyzing a SWE-bench Verified coding task to predict its difficulty. This is a BUG FIX task in a Python repository. You have access to the full task information including the gold solution patch.
\end{promptbox}

\begin{promptbox}[title=Introduction --- SWE-bench Pro]
You are analyzing a SWE-bench Pro coding task to predict its difficulty. This is a BUG FIX task in a Python repository. SWE-bench Pro contains more challenging tasks than standard SWE-bench. You have access to the full task information including the gold solution patch.
\end{promptbox}

\begin{promptbox}[title=Introduction --- Terminal-Bench 2.0]
You are analyzing a TerminalBench terminal/shell task to predict its difficulty. This task requires writing shell commands or scripts to accomplish a goal. You have access to the full task information including the reference solution.
\end{promptbox}

\begin{promptbox}[title=Introduction --- GSO]
You are analyzing a GSO (Software Optimization Benchmark) task to predict its difficulty. This is a PERFORMANCE OPTIMIZATION task, NOT a bug fix. The goal is to make code run faster while maintaining correctness. You have access to the full task information including the gold optimization patch.
\end{promptbox}

After the introduction, the following completeness instruction is included:

\begin{promptbox}[title=Completeness Instruction]
CRITICAL: You MUST provide a value for EVERY feature listed below. Do not skip any features. If uncertain, provide your best estimate. Missing values will cause extraction to fail.
\end{promptbox}

The \textbf{task information} block varies by dataset. We show the f-string template for each dataset with all feature sources available in the prompt. Patches are truncated to 300K characters, test patches to 200K characters, and regression test lists to 50K characters.

\begin{promptbox}[title=Task Information Template --- SWE-bench Verified / Pro]
## TASK INFORMATION

**Instance ID:** {instance_id}
**Repository:** {repo}
**Version:** {version}

**Problem Statement:**
{problem_statement}

**Gold Patch (correct solution):**
```diff
{patch}
```

**Test Patch (tests that verify the fix):**
```diff
{test_patch}
```

**Tests that should pass after fix (FAIL_TO_PASS):**
{fail_to_pass}

**Regression tests (PASS_TO_PASS):**
{pass_to_pass}
\end{promptbox}
The ``Claimed Difficulty" field for Terminal-Bench 2.0 below is simply a label of ``easy," ``medium," or ``hard,"  present in the dataset itself and not related to the IRT difficulties. We left this in the prompt because we felt that in our main use case of benchmark designing, it is reasonable that the designer has some prior about the difficulty of the task at this broad level.
\begin{promptbox}[title=Task Information Template --- Terminal-Bench 2.0]
## TASK INFORMATION

**Task ID:** {task_id}
**Category:** {category}
**Tags:** {tags}
**Claimed Difficulty:** {difficulty}

**Task Instruction:**
{problem_statement}

**Evaluation Test Harness:**
```
{tests}
```

**Reference Solution (solution.sh):**
```bash
{patch}
```
\end{promptbox}

\begin{promptbox}[title=Task Information Template --- GSO]
## TASK INFORMATION

**Instance ID:** {instance_id}
**Repository:** {repo}
**API/Function being optimized:** {api}

**Performance Benchmark Script:**
```python
{prob_script}
```

**Gold Patch (optimization solution):**
```diff
{gt_diff}
```
\end{promptbox}

Below are the scale definitions for all 12 non-environment features using the SWE-bench variant. Per-dataset scale variants for Terminal-Bench~2.0 and GSO are shown in Table~\ref{tab:scale-variants}.

\begin{promptbox}[title=Feature Scales --- SWE-bench (12 Non-Environment Features)]
## FEATURES TO EVALUATE

Analyze the task information above to evaluate these features. Be precise and consistent with your ratings.

### Solution Hint (solution_hint: 0-3)
Does the task description contain or hint at the solution approach?
- 0: No hint at the solution at all
- 1: Vague hint or general direction
- 2: Clear description of approach needed
- 3: Exact solution or detailed steps provided

### Domain Knowledge Required (domain_knowledge_required: 1-5)
How much specialized knowledge is needed?
- 1: Basic Python, obvious fix anyone could make
- 2: Standard library knowledge needed
- 3: Framework-specific knowledge (Django, pytest, numpy, etc.)
- 4: Deep understanding of the library's internals
- 5: Obscure APIs, protocols, or highly specialized domain knowledge

### Logical Reasoning Required (logical_reasoning_required: 1-5)
How much logical reasoning is needed?
- 1: Mechanical execution, no reasoning needed
- 2: Simple cause-effect reasoning
- 3: Multi-step reasoning required
- 4: Complex reasoning with multiple factors
- 5: Deep reasoning about edge cases, invariants, or system behavior

### Atypicality (atypicality: 1-5)
How unusual is this task pattern?
- 1: Very common pattern (well-known solution approach)
- 2: Common pattern in this domain
- 3: Moderately unusual
- 4: Unusual pattern
- 5: Rare or novel pattern

### Verification Difficulty (verification_difficulty: 1-5)
How hard is it to verify the solution is correct?
- 1: Trivial (obvious pass/fail)
- 2: Easy (straightforward test cases)
- 3: Moderate (some edge cases to consider)
- 4: Hard (subtle correctness issues, complex setup)
- 5: Very hard (rare edge cases, hard to reproduce, timing-sensitive)

### Error Specificity (error_specificity: 1-5)
How specific is the error or bug description?
- 1: Very vague symptoms, unclear what's actually broken
- 2: General description of misbehavior
- 3: Specific behavior described but no error details
- 4: Clear error description with some context
- 5: Exact error message, stack trace, or precise failure mode

### Debugging Complexity (debugging_complexity: 1-5)
Based on the problem description, how complex would root cause analysis be?
- 1: Obvious cause stated or implied
- 2: Straightforward to identify cause
- 3: Moderate investigation needed
- 4: Complex debugging likely required
- 5: Deep investigation into internals needed

### Codebase Scope (codebase_scope: 1-5)
How much of the codebase might need to be understood or modified?
- 1: Likely isolated to single file/function
- 2: Few related files
- 3: Multiple components involved
- 4: Cross-cutting concern affecting many areas
- 5: System-wide implications

### Similar Issue Likelihood (similar_issue_likelihood: 0/1)
Is this likely a common bug type that has known solutions?
- 0: Novel or unusual issue, unlikely to find similar cases
- 1: Common bug pattern (e.g., null check, encoding, off-by-one, race condition)

### Side Effect Risk (side_effect_risk: 1-5)
How likely are unintended side effects from the fix?
- 1: No risk (internal change, no API implications)
- 2: Minor API implications
- 3: Some compatibility or behavioral side effects possible
- 4: Significant API/behavior changes likely
- 5: Critical risk -- backwards compatibility, deprecation, or wide-reaching behavioral changes

### Test Edge Case Coverage (test_edge_case_coverage: 1-5)
Does the test cover edge cases, boundary conditions, and error scenarios?
- 1: Happy path only - no edge cases tested
- 2: Minimal - one or two edge cases
- 3: Moderate - some boundary conditions checked
- 4: Good - most edge cases and error conditions tested
- 5: Thorough - comprehensive edge case and error handling coverage

### Solution Complexity (solution_complexity: 1-5)
How complex is the actual code change?
- 1: Trivial (add parameter, change value, simple one-liner)
- 2: Simple (straightforward logic change)
- 3: Moderate (requires understanding context, multiple changes)
- 4: Complex (algorithmic changes, multiple interdependent fixes)
- 5: Very complex (architectural changes, subtle edge cases)
\end{promptbox}

Each batch request ends with an \textbf{output format} specification:

\begin{promptbox}[title=Output Format]
## OUTPUT FORMAT

Respond with ONLY a JSON object containing your ratings. Do not include any explanation or commentary outside the JSON.

```json
{"feature_1": <min-max>, "feature_2": <min-max>, ...}
```
\end{promptbox}

$7$ of the $12$ features have per-dataset scale text variants. Table~\ref{tab:scale-variants} shows the Terminal-Bench~2.0 and GSO variants; the SWE-bench variant is shown above.

\begin{table}[h]
\scriptsize
\centering
\begin{tabularx}{\textwidth}{c X X}
\toprule
& {\small\textbf{Terminal-Bench 2.0}} & {\small\textbf{GSO}} \\
\midrule
\multicolumn{3}{l}{\small\textbf{domain\_knowledge\_required} (1--5)} \\
1 & Basic shell commands anyone could use (ls, cd, cat, echo) & Basic Python performance (list comprehensions, generators) \\
2 & Standard Unix tools (grep, sed, awk, find) & Standard library optimization patterns \\
3 & Specialized tools or configurations (cmake, git internals, network tools) & Library-specific knowledge (numpy, pandas internals) \\
4 & Deep understanding of systems (kernel, filesystems, protocols) & Deep understanding of library implementation \\
5 & Obscure tools, APIs, or highly specialized domain knowledge & Expert knowledge (SIMD, memory layout, CPU caches) \\
\midrule
\multicolumn{3}{l}{\small\textbf{error\_specificity} (1--5)} \\
1 & Very vague goal, unclear what success looks like & Vague ``make it faster'' with no specifics \\
2 & General description of desired outcome & General description of slowness \\
3 & Specific outcome described but details missing & Specific function/API identified as slow \\
4 & Clear description with context about expected behavior & Clear performance issue with some context \\
5 & Exact specification with precise success criteria & Exact bottleneck identified with profiling data or benchmarks \\
\midrule
\multicolumn{3}{l}{\small\textbf{debugging\_complexity} (1--5)} \\
1 & Obvious approach stated or implied & Obvious bottleneck stated or implied \\
2 & Straightforward to determine approach & Straightforward to profile and identify \\
3 & Moderate exploration/research needed & Moderate profiling/analysis needed \\
4 & Complex problem-solving likely required & Complex performance analysis likely required \\
5 & Deep investigation into tools/systems needed & Deep investigation into runtime behavior needed \\
\midrule
\multicolumn{3}{l}{\small\textbf{similar\_issue\_likelihood} (0/1)} \\
0 & Novel or unusual task, unlikely to find similar examples online & Novel bottleneck requiring creative optimization strategy \\
1 & Common pattern (e.g., file processing, service configuration, text extraction) & Common optimization pattern (e.g., vectorization, caching, batch processing, algorithmic improvement) \\
\midrule
\multicolumn{3}{l}{\small\textbf{side\_effect\_risk} (1--5)} \\
1 & No risk (self-contained operation, no system state changes) & No risk (simple speedup, identical behavior guaranteed) \\
2 & Minor filesystem or config changes & Minor numerical precision differences possible \\
3 & Some risk of affecting other services or system state & Some edge cases might behave differently \\
4 & Significant risk of breaking other processes or configurations & Significant behavioral changes in corner cases likely \\
5 & Critical risk --- system-wide changes, network/security implications & Critical risk --- optimization fundamentally changes semantics or data flow \\
\midrule
\multicolumn{3}{l}{\small\textbf{test\_edge\_case\_coverage} (1--5)} \\
1 & Happy path only --- no edge cases tested & Happy path only --- typical input sizes only \\
2 & Minimal --- one or two edge cases & Minimal --- one or two boundary sizes \\
3 & Moderate --- some boundary conditions checked & Moderate --- some degenerate inputs (empty, very large) \\
4 & Good --- most edge cases and failure modes tested & Good --- includes unusual shapes, dtypes, memory layouts, special values \\
5 & Thorough --- comprehensive edge case, error handling, and adversarial input coverage & Thorough --- comprehensive degenerate cases, adversarial inputs, and stress tests \\
\midrule
\multicolumn{3}{l}{\small\textbf{solution\_complexity} (1--5)} \\
1 & Trivial (single command, simple file operation) & Simple (add caching, use built-in function) \\
2 & Simple (straightforward multi-step task) & Standard (vectorization, batch processing) \\
3 & Moderate (requires understanding context, multiple tools) & Moderate (algorithm improvements, memory optimization) \\
4 & Complex (multiple interdependent steps, debugging needed) & Complex (significant algorithmic changes) \\
5 & Very complex (multi-stage pipeline, cross-system integration) & Very complex (architectural redesign, low-level optimization) \\
\bottomrule
\end{tabularx}
\caption{Per-dataset scale text variants for features that differ across datasets. Five features (\texttt{solution\_hint}, \texttt{logical\_reasoning\_required}, \texttt{atypicality}, \texttt{verification\_difficulty}, and \texttt{codebase\_scope}) use the same scale text across all datasets and are omitted.}
\label{tab:scale-variants}
\end{table}

\subsubsection{Auditor Agent Prompt}
\label{app:auditor-prompt}

The auditor agent extracts 8 environment features, of which 3 (\texttt{codebase\_scale}, \texttt{fix\_localization}, and \texttt{implementation\_language\_complexity}) are used in the default feature set (i.e. for all experiments besides the feature source ablation). Below is the full prompt (shown for SWE-bench Verified; the task context paragraph varies by dataset). We note that for GSO, for the auditor agent used in the non feature source ablation experiments, we gave it the test case (i.e. benchmarking script) as well, as we felt that having the name of the function to be optimized only was too restrictive,  and we should have this affordance anyways since we are making the test case and solution available in the prompts for all the non-environment features. 

\begin{promptbox}[title=Auditor Agent Prompt (SWE-bench Verified)]
You are a codebase auditor evaluating task environments for difficulty prediction. Your job is to explore the environment and rate it on 8 difficulty-related axes.

## Task Context

You are auditing a **bug-fix** task in a Python repository. The /testbed directory contains the project codebase with a known bug. The problem statement describes the bug, and there are failing tests (FAIL_TO_PASS) that should pass after the fix.

## Your Task

1. Explore the working directory to understand the project structure
2. Read the problem statement (provided as input)
3. Try to understand the scope and complexity of the task
4. Check available tools, tests, and dependencies
5. Rate the environment on the 8 axes below

## Features to Assess (1-5 scale)

### Fix Localization (fix_localization: 1-5)
How spread out is the likely solution?
- 1: Solution requires changes across many modules/packages
- 2: Solution spans multiple files across different directories
- 3: Solution spans 2-3 files in the same module
- 4: Solution is in 1-2 closely related files
- 5: Solution is contained to a single function/method

### Entry Point Clarity (entry_point_clarity: 1-5)
How easy is it to find where the problem manifests?
- 1: No clear entry point, requires deep architecture knowledge
- 2: Entry point exists but buried in abstraction layers
- 3: Entry point findable with moderate searching
- 4: Problem statement or tests hint at the location
- 5: Clear from problem statement exactly which file/function

### Change Blast Radius (change_blast_radius: 1-5)
How many components would be affected by changes? (Higher = harder)
- 1: Isolated change, no downstream effects
- 2: Minor coupling, 1-2 related files to consider
- 3: Moderate coupling, changes affect a subsystem
- 4: High coupling, changes ripple across modules
- 5: Core/shared code, changes affect entire codebase

### Environment Setup Complexity (environment_setup_complexity: 1-5)
How complex is the runtime/tooling environment?
- 1: Standard single-directory project, ready to run out of the box
- 2: Minor configuration needed, clear project structure
- 3: Multiple services or components, custom configurations
- 4: Complex orchestration, specialized dependencies, non-trivial build steps
- 5: Exotic environment, multi-container setup, hardware-specific requirements

### Implementation Language Complexity (implementation_language_complexity: 1-5)
How complex is the primary language/tech stack for the solution?
- 1: Pure Python or simple shell commands
- 2: Python with standard libraries, basic scripting
- 3: Mixed languages (Python + build tools), moderately complex shell
- 4: Compiled languages (C/C++), complex build systems, framework-specific patterns
- 5: Multi-language (C/Rust + Python bindings), SIMD/assembly, exotic toolchains

### Testing Infrastructure Quality (testing_infrastructure_quality: 1-5)
How good is the testing/validation setup for verifying a solution?
- 1: No test framework, no way to validate changes
- 2: Basic tests exist but hard to run or incomplete
- 3: Standard test framework, moderate coverage
- 4: Good test coverage, easy to run tests, clear pass/fail signals
- 5: Comprehensive test suite, fast feedback loops, detailed error messages

### Dependency Complexity (dependency_complexity: 1-5)
How complex are the project dependencies?
- 1: No external dependencies, standard library only
- 2: Few well-known dependencies (e.g., requests, numpy)
- 3: Moderate number of standard packages
- 4: Many dependencies, some specialized or version-sensitive
- 5: Complex dependency tree, C extensions, system-level deps, version conflicts

### Codebase Scale (codebase_scale: 1-5)
How large/complex is the codebase the agent needs to work with?
- 1: Tiny project (<100 files, <5K lines)
- 2: Small project (100-500 files)
- 3: Medium project (500-2000 files)
- 4: Large project (2000-10000 files)
- 5: Massive project (10000+ files, complex module structure)

## Output Format

After your exploration (use 8-15 tool calls), output your final assessment as a JSON object with exactly 8 features. Each feature should be an object with "value" (1-5 integer) and "reasoning" (brief explanation):

```json
{
  "fix_localization": {"value": 3, "reasoning": "Brief explanation"},
  "entry_point_clarity": {"value": 3, "reasoning": "Brief explanation"},
  "change_blast_radius": {"value": 3, "reasoning": "Brief explanation"},
  "environment_setup_complexity": {"value": 3, "reasoning": "Brief explanation"},
  "implementation_language_complexity": {"value": 3, "reasoning": "Brief explanation"},
  "testing_infrastructure_quality": {"value": 3, "reasoning": "Brief explanation"},
  "dependency_complexity": {"value": 3, "reasoning": "Brief explanation"},
  "codebase_scale": {"value": 3, "reasoning": "Brief explanation"}
}
```

**CRITICAL**: Your final message MUST contain a valid JSON object with all 8 features. Do not forget any features.

## Tips

- Start with `ls` or `find` to understand the project structure
- Check for test files, configuration files, and dependency files
- Look at file extensions to understand the tech stack
- Use `wc -l` or `find . -type f | wc -l` to gauge codebase size
- Check `requirements.txt`, `setup.py`, `Cargo.toml`, etc. for dependencies
- Keep your exploration focused - aim for 8-15 tool calls

## IMPORTANT: How to Complete

After 8-15 exploration commands, you MUST call the `submit()` function with your JSON report.
Do NOT try to solve the task - just audit and rate the environment.

Now begin your audit. Start by exploring the working directory structure, then submit your ratings.
\end{promptbox}

\subsubsection{Feature Source Ablation Prompts}
\label{app:info-ablation-prompts}

For the feature source ablation experiment (Table \ref{tab:information-ablation}), we extract the full set of 28 features at each information level. Beyond the 12 features in the main experiments and the 8 environment features extracted by the auditor agent, we extract 8 additional features. These additional features are used only in the information ablation, where the top 15 features are selected per information level. Below are their scale definitions (SWE-bench variant).

\begin{promptbox}[title=Additional Problem Features (5)]
### Problem Clarity (problem_clarity: 1-5)
How clear and well-specified is the task?
- 1: Very vague, unclear what's actually required
- 2: Somewhat clear but missing key details
- 3: Reasonably clear, some ambiguity
- 4: Clear with good context
- 5: Crystal clear with explicit steps and expected behavior

### Standard Pattern Available (standard_pattern_available: 0/1)
Is this a well-documented pattern with existing examples?
- 0: Novel solution needed, no clear pattern to follow
- 1: Well-documented pattern (e.g., common idiom, StackOverflow answer available)

### Reproduction Clarity (reproduction_clarity: 1-5)
How clear are the steps to reproduce the issue?
- 1: No reproduction steps, unclear how to trigger
- 2: Vague conditions mentioned
- 3: General scenario described
- 4: Clear steps but some setup unclear
- 5: Exact reproduction steps with code/commands provided

### Expected Behavior Clarity (expected_behavior_clarity: 1-5)
How clear is what the correct behavior should be?
- 1: Very ambiguous, multiple interpretations possible
- 2: General expectation but details unclear
- 3: Reasonably clear expected outcome
- 4: Clear expected behavior with examples
- 5: Precisely specified with exact expected output/behavior

### Information Completeness (information_completeness: 1-5)
How complete is the information provided in the problem statement?
- 1: Missing critical information, many unknowns
- 2: Key details missing
- 3: Adequate information but gaps exist
- 4: Good context provided
- 5: Comprehensive information including versions, configs, examples
\end{promptbox}

\begin{promptbox}[title=Additional Test Features (2)]
### Test Comprehensiveness (test_comprehensiveness: 1-5)
How thoroughly does the test patch cover the expected behavior?
- 1: Minimal - tests only one basic case
- 2: Limited - tests a few cases but misses important scenarios
- 3: Moderate - covers main functionality with some gaps
- 4: Good - covers most expected behaviors and variations
- 5: Exhaustive - comprehensive coverage including corner cases

### Test Assertion Complexity (test_assertion_complexity: 1-5)
How complex are the assertions and test setup in the test patch?
- 1: Simple - basic equality checks (assertEqual, assertTrue)
- 2: Standard - uses common assertion patterns
- 3: Moderate - multiple assertions, some setup required
- 4: Complex - requires mocking, fixtures, or intricate setup
- 5: Very complex - extensive mocking, async testing, or multi-step verification
\end{promptbox}

\begin{promptbox}[title=Additional Solution Feature (1)]
### Integration Complexity (integration_complexity: 1-5)
How tightly must the changes integrate with existing code?
- 1: Self-contained/greenfield - new code with clear boundaries
- 2: Simple extension - adds to existing code with clear interface
- 3: Moderate integration - changes interact with several existing components
- 4: Deep integration - requires understanding multiple subsystems
- 5: Pervasive integration - affects system-wide behavior, many touchpoints
\end{promptbox}

Five of these additional features also have per-dataset scale variants:
\begin{itemize}
    \item \texttt{reproduction\_clarity},
    \item \texttt{expected\_behavior\_clarity},
    \item \texttt{test\_comprehensiveness},
    \item \texttt{test\_assertion\_complexity},
    \item \texttt{integration\_complexity}.
\end{itemize}
Table~\ref{tab:additional-scale-variants} shows the Terminal-Bench~2.0 and GSO variants.

\begin{table}[h]
\scriptsize
\centering
\begin{tabularx}{\textwidth}{c X X}
\toprule
& {\small\textbf{Terminal-Bench 2.0}} & {\small\textbf{GSO}} \\
\midrule
\multicolumn{3}{l}{\small\textbf{reproduction\_clarity} (1--5)} \\
1 & No setup steps, unclear how to begin & No benchmark or test scenario provided \\
2 & Vague environment requirements mentioned & Vague description of slow use case \\
3 & General setup described & General performance scenario described \\
4 & Clear steps but some prerequisites unclear & Clear benchmark but some parameters unclear \\
5 & Exact setup and execution steps with commands provided & Exact benchmark with input sizes and expected speedup \\
\midrule
\multicolumn{3}{l}{\small\textbf{expected\_behavior\_clarity} (1--5)} \\
1 & Very ambiguous, multiple valid interpretations & Very ambiguous, unclear what ``faster'' means in context \\
2 & General goal but details unclear & General speedup goal but no specifics \\
3 & Reasonably clear target outcome & Target function/API clear but speedup threshold unclear \\
4 & Clear success criteria with examples & Clear optimization target with approximate goals \\
5 & Precisely specified with exact expected output/state & Precisely specified with exact performance requirements \\
\midrule
\multicolumn{3}{l}{\small\textbf{test\_comprehensiveness} (1--5)} \\
1 & Minimal --- checks only one basic output & Minimal --- tests only one basic case with trivial input \\
2 & Limited --- checks a few conditions but misses important scenarios & Limited --- tests a few input sizes but misses important scenarios \\
3 & Moderate --- covers main success criteria with some gaps & Moderate --- covers main use case with some size variations \\
4 & Good --- covers most expected outcomes and edge cases & Good --- covers multiple input sizes, shapes, and data types \\
5 & Exhaustive --- comprehensive coverage including corner cases and error handling & Exhaustive --- comprehensive coverage including edge cases and realistic workloads \\
\midrule
\multicolumn{3}{l}{\small\textbf{test\_assertion\_complexity} (1--5)} \\
1 & Simple --- basic file existence or string match check & Simple --- basic equality check (reference.equals(current)) \\
2 & Standard --- checks output format or simple numeric comparison & Standard --- checks a few output fields individually \\
3 & Moderate --- multiple checks, some parsing of output required & Moderate --- multiple assertions, type/shape checking, tolerance-based comparison \\
4 & Complex --- statistical validation, multi-step verification, or custom scoring & Complex --- custom equivalence logic, statistical validation, or multi-step verification \\
5 & Very complex --- cross-referencing multiple outputs, timing-sensitive checks & Very complex --- domain-specific correctness checks, numerical stability verification \\
\midrule
\multicolumn{3}{l}{\small\textbf{integration\_complexity} (1--5)} \\
1 & No special tools needed (basic shell) & Self-contained optimization with clear boundaries \\
2 & Standard development tools (git, make, pip) & Simple drop-in replacement for existing function \\
3 & Multiple specialized tools or complex configuration & Moderate integration --- optimization touches several components \\
4 & Uncommon tools or complex build systems & Deep integration --- requires understanding data flow across subsystems \\
5 & Exotic toolchain, legacy systems, or cross-compilation & Pervasive changes --- optimization affects system-wide architecture \\
\bottomrule
\end{tabularx}
\caption{Per-dataset scale text variants for the 5 additional features (from the information ablation) that differ across datasets.}
\label{tab:additional-scale-variants}
\end{table}

In the information ablation, the introduction is adjusted to reflect what information is available. We show the SWE-bench Verified variants at the problem and test levels (the solution level is shown above):

\begin{promptbox}[title=Introduction --- SWE-bench Verified (problem level)]
You are analyzing a SWE-bench Verified coding task to predict its difficulty. This is a BUG FIX task in a Python repository. You only have access to the problem statement.
\end{promptbox}

\begin{promptbox}[title=Introduction --- SWE-bench Verified (test level)]
You are analyzing a SWE-bench Verified coding task to predict its difficulty. This is a BUG FIX task in a Python repository. You have access to the problem statement and the test patch, but NOT the solution.
\end{promptbox}

The task information template also varies by level. We show the SWE-bench templates at the problem and test levels:

\begin{promptbox}[title=Task Information Template --- SWE-bench (problem level)]
## TASK INFORMATION

**Instance ID:** {instance_id}

**Problem Statement:**
{problem_statement}
\end{promptbox}

\begin{promptbox}[title=Task Information Template --- SWE-bench (test level)]
## TASK INFORMATION

**Instance ID:** {instance_id}
**Repository:** {repo}
**Version:** {version}

**Problem Statement:**
{problem_statement}

**Test Patch (tests that verify the fix):**
```diff
{test_patch}
```

**Tests that should pass after fix (FAIL_TO_PASS):**
{fail_to_pass}
\end{promptbox}

\section{Validation of the LLM-Scaffold Decomposition of Agent Ability}
\label{app:validation-of-decomposition}

\subsection{Full Results of the New Responses Experiment}
\label{app:new_responses}
In Table \ref{tab:new-responses}, we present the results of the \textit{New Responses} experiment described in Section \ref{sec:new-responses}.

\begin{table*}[h]
\small
\centering
\begin{tabular}{lcc:c}
    \toprule
    Benchmark & IRT-Agent & Standard IRT & Oracle \\
    \midrule
    SWE-bench Verified & 0.939 $\pm$ 0.000 & 0.940 $\pm$ 0.000 & 0.945 $\pm$ 0.000 \\
    SWE-bench Pro & 0.870 $\pm$ 0.002 & 0.875 $\pm$ 0.002 & 0.919 $\pm$ 0.000 \\
    GSO & 0.810 $\pm$ 0.014 & 0.814 $\pm$ 0.015 & 0.920 $\pm$ 0.001 \\
    Terminal-Bench 2.0 & 0.925 $\pm$ 0.000 & 0.925 $\pm$ 0.001 & 0.935 $\pm$ 0.000 \\
    \midrule
    \textit{All benchmarks} & 0.930 $\pm$ 0.000 & --- & --- \\
    \bottomrule
\end{tabular}
\caption{\textbf{AUC-ROC (mean $\pm$ std of cross-validation means across 20 random seeds) on held-out responses}. \textit{IRT-Agent}---our proposed IRT model with agent features---performs on par with \textit{Standard IRT} despite being strictly less expressive, while enabling multi-benchmark training and predictions with high performance. Both approach \textit{Oracle}, which is another standard IRT model trained on all data, including the held-out responses.}
\label{tab:new-responses}
\end{table*}

\subsection{Qualitative Inspection of Learned Parameters in a Multi-Benchmark IRT with Agent Features}
\label{app:qualitative-inspection}

\begin{figure}[h]
    \centering
    \includegraphics[width=0.7\linewidth]{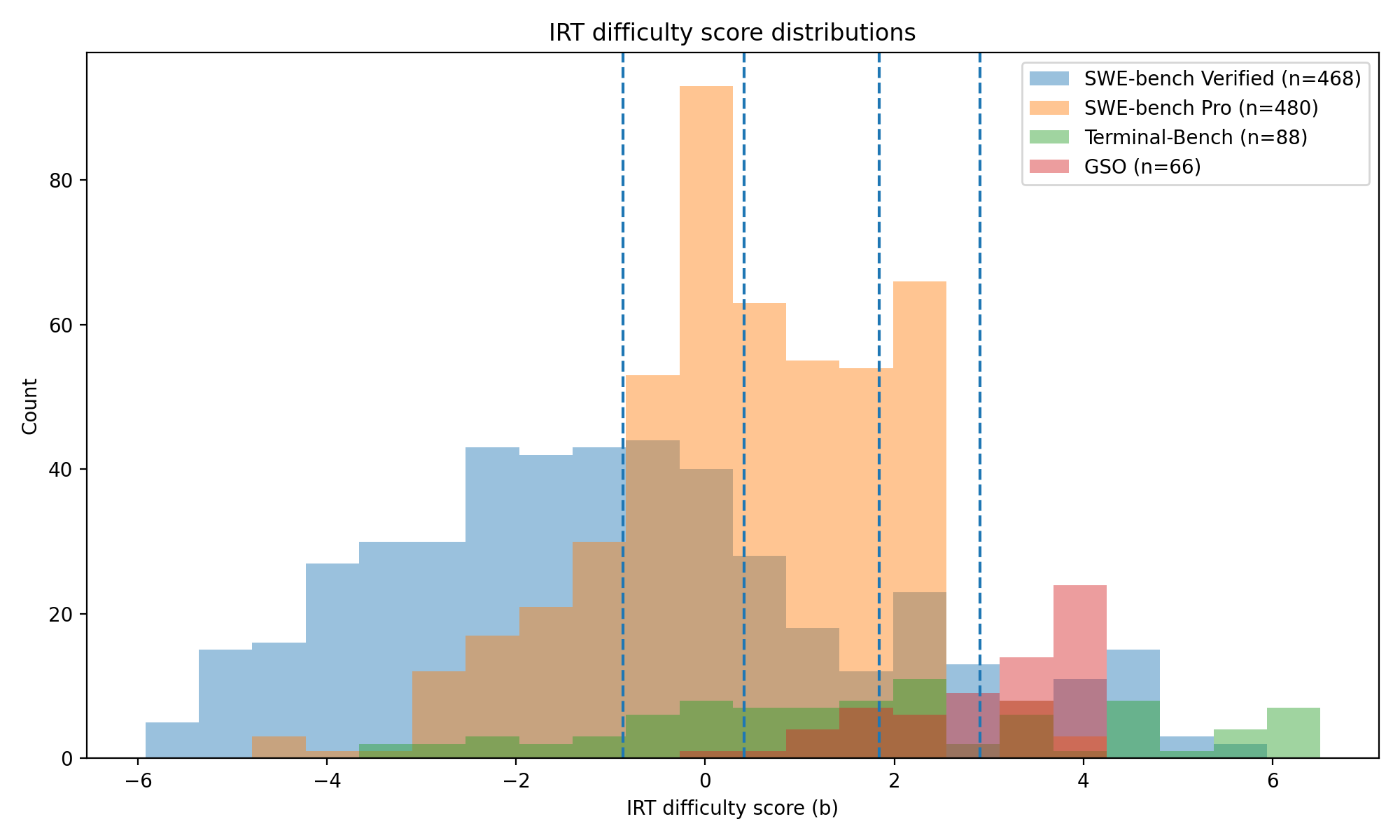}
    \caption{\textbf{Task difficulty histograms.} SWE-bench Pro is harder on average than Verified, GSO is the hardest, and Terminal-Bench 2.0 is highly heterogeneous.}
    \label{fig:difficulty-histogram}
\end{figure}

We train a four-benchmark IRT with agent features on all data except zero-solve tasks in order to compare task difficulties across benchmarks and see if they qualitatively match our expectations. Figure \ref{fig:difficulty-histogram} shows the result: SWE-bench Pro (mean $b=0.407$) is harder on average than SWE-bench Verified (mean $b=-0.867$), as intended by the developers of the former \citep{deng2025swebenchproaiagents}. GSO (mean $b=2.907$) is the hardest of all four benchmarks, consistent with the finding of \citet{ho2025rosetta}, who identified GSO as one of the hardest benchmarks overall.

We additionally look at the learned LLM and scaffold ability parameters.

\begin{table}[h]
\centering
\small
\begin{tabular}{clc}
\toprule
Rank & LLM & Ability ($\theta_m$) \\
\midrule
1 & Gemini 3.1 Pro & 3.660 \\
2 & GPT-5.3-Codex & 3.440 \\
3 & Claude Opus 4.6 & 3.332 \\
4 & GPT-5.2-Codex & 2.997 \\
5 & GPT-5.2 & 2.697 \\
6 & Claude Opus 4.5 & 2.607 \\
7 & Gemini 3 Pro & 2.493 \\
8 & \texttt{GLM-5} & 2.461 \\
9 & GPT-5.1-Codex-Max & 2.429 \\
10 & Gemini 3 Flash & 2.094 \\
11 & GPT-5 & 2.014 \\
12 & Claude Sonnet 4.5 & 2.012 \\
13 & GPT-5.1-Codex & 1.894 \\
14 & \texttt{Kimi K2.5} & 1.699 \\
15 & \texttt{MiniMax-M2.5} & 1.624 \\
\midrule
60 & \texttt{Skywork-SWE-32B} & -1.298 \\
61 & Claude 3.5 Haiku & -1.658 \\
62 & \texttt{Lingma-SWE-GPT-72B} & -1.760 \\
63 & \texttt{Qwen2.5-72B} & -1.928 \\
64 & GPT-5-Nano & -2.211 \\
65 & GPT-4 & -2.597 \\
66 & \texttt{MCTS-Refine-7B} & -2.600 \\
67 & GPT-4o & -2.714 \\
68 & Claude 3 Opus & -2.815 \\
69 & Claude 2 & -2.859 \\
70 & \texttt{Lingma-SWE-GPT-7B} & -3.073 \\
71 & \texttt{GPT OSS 20B} & -4.025 \\
72 & \texttt{SWE-Llama-7B} & -4.047 \\
73 & \texttt{SWE-Llama-13B} & -4.204 \\
74 & GPT-3.5 & -5.014 \\
\bottomrule
\end{tabular}
\caption{\textbf{Top-15 and bottom-15 LLMs by IRT ability.} \texttt{Teletype font} indicates open-weight LLMs.}
\label{tab:llm-abilities}
\end{table}

The LLM abilities (Table \ref{tab:llm-abilities}) make sense: the top-3 LLMs are Gemini 3.1 Pro, GPT-5.3-Codex, and Claude Opus 4.6, which were widely regarded to be the most capable coding LLMs at the time of the experiment, while the LLM with the lowest ability value is GPT-3.5 (the oldest LLM included in the leaderboards).

For many scaffolds, however, the ability parameter cannot be accurately calibrated because the scaffold occurs only once in the data (46 out of 72 scaffolds in this analysis). Scaffolds occurring once with a capable LLM tend to have high ability values, while common scaffolds or those occurring once with a less-capable LLM have lower values (Table \ref{tab:scaffold-abilities}). Claude Code has a surprisingly low ability value, but this is supported by the data: on Terminal-Bench 2.0 (the only benchmark on which Claude Code is evaluated), the same LLM (Claude Opus 4.5) performs better with multiple other scaffolds (Factory Droid, Letta Code, Terminus 2, Goose) than with Claude Code.

\begin{table}[h]
\centering
\small
\begin{tabular}{clc}
\toprule
Rank & Scaffold & Ability ($\theta_s$) \\
\midrule
1 & \texttt{AgentScope} & 1.453 \\
2 & Learn-by-interact & 1.207 \\
3 & Ante & 1.060 \\
4 & Junie & 0.919 \\
5 & \texttt{AutoCodeRover v20240620} & 0.896 \\
6 & Z.ai & 0.805 \\
7 & \texttt{ForgeCode} & 0.769 \\
8 & \texttt{Terminus KIRA} & 0.729 \\
9 & \texttt{Agentless 1.5} & 0.715 \\
10 & Harness AI & 0.664 \\
11 & Devlo & 0.652 \\
12 & \texttt{Lingxi v1.5} & 0.611 \\
13 & Simple Codex & 0.581 \\
14 & \texttt{OpenHands CodeAct 2.1} & 0.545 \\
15 & MASAI & 0.513 \\
\midrule
58 & \texttt{mini-SWE-agent} & -0.463 \\
59 & \texttt{Terminus 2} & -0.490 \\
60 & \texttt{Goose} & -0.545 \\
61 & \texttt{OpenCode} & -0.555 \\
62 & \texttt{R2E-Gym} & -0.562 \\
63 & MAYA & -0.643 \\
64 & \texttt{SWE-agent} & -0.706 \\
65 & Claude Code & -0.713 \\
66 & Dakou Agent & -0.820 \\
67 & \texttt{SWE-agent 1.0} & -0.823 \\
68 & SWE-Rizzo & -0.865 \\
69 & LingmaAgent & -0.938 \\
70 & Artemis Agent v2 & -1.053 \\
71 & Artemis Agent v1 & -1.133 \\
72 & \texttt{RAG} & -2.954 \\
\bottomrule
\end{tabular}
\caption{\textbf{Top-15 and bottom-15 scaffolds by IRT ability.} \texttt{Teletype font} indicates open-source scaffolds.}
\label{tab:scaffold-abilities}
\end{table}

\subsection{Comparison of LLM Abilities in Multi-Scaffold and Fixed-Scaffold Settings}
\label{app:terminus}
\begin{figure}[h]
    \centering
    \includegraphics[width=0.75\linewidth]{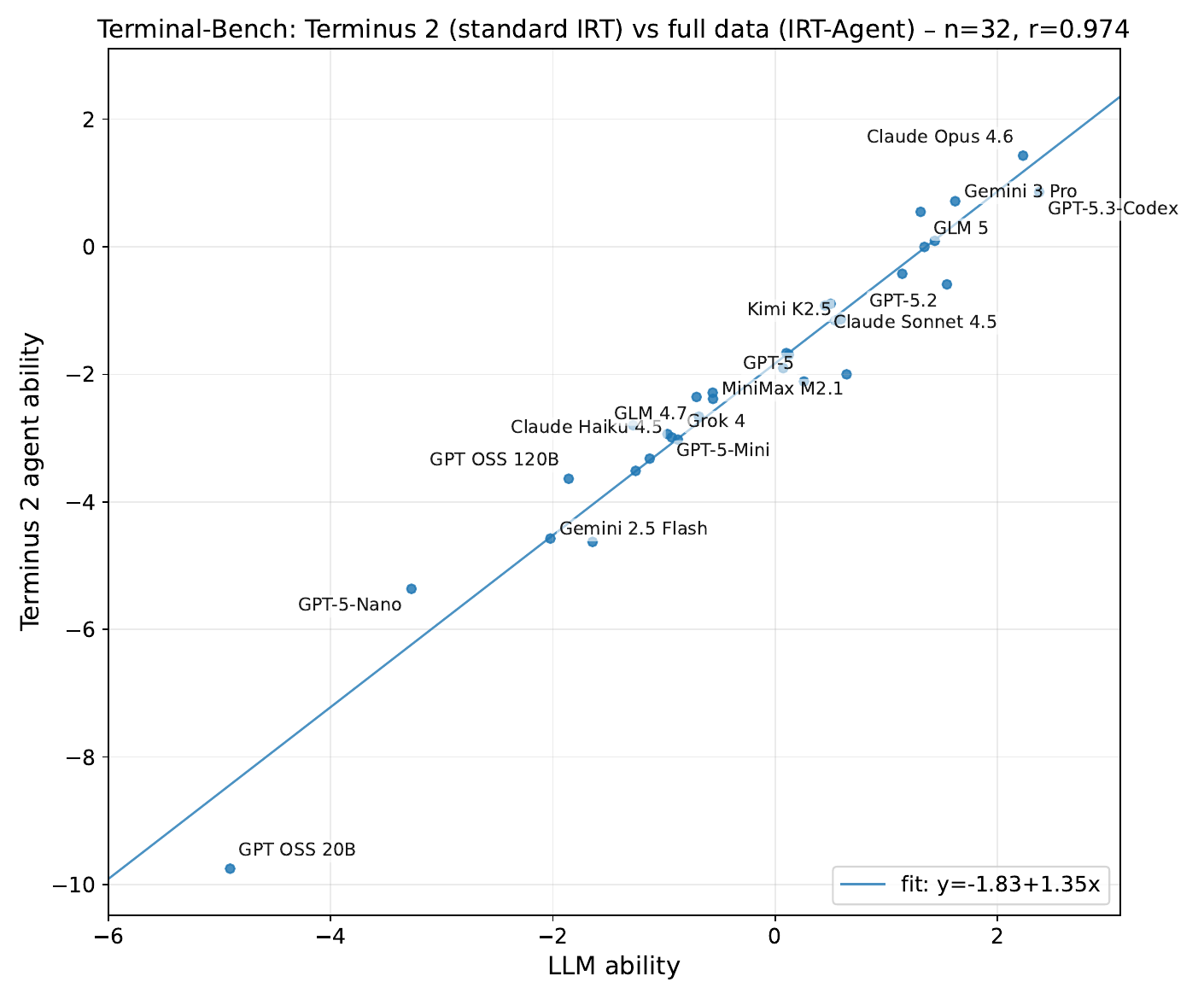}
    \caption{\textbf{Validation of decomposition.} Strong correlation (Pearson $r=0.974$) between agent abilities learned on a fixed scaffold (Terminus 2) versus LLM abilities isolated via our decomposition method.}
    \label{fig:terminalbench-scatterplot}
\end{figure}

We leverage Terminal-Bench 2.0, whose authors evaluate all LLMs with the same fixed scaffold Terminus 2, in addition to third-party evaluations that can use any scaffold \citep{merrill2026terminalbench}. Hence, if we train a standard IRT on the subset of responses with the Terminus 2 scaffold, the ability of each agent should be roughly the same as the ability of the corresponding LLM if we train IRT with agent features on the full benchmark data. In Figure \ref{fig:terminalbench-scatterplot}, we plot the agent abilities from the Terminus 2 subset against the LLM abilities from the full data trained using our model-scaffold decomposition. We see a strong correlation between the two, validating our interpretation of the LLM ability parameter. We note that the full response set has 38 unique LLMs and 112 LLM-scaffold combinations, for an average of 2.98 scaffolds per LLM. That is, the average LLM is evaluated with Terminus 2 plus two other scaffolds. This relatively small number of different scaffolds per LLM may explain why the LLM-scaffold decomposition does not significantly constrain the expressivity of the proposed IRT model with agent features, resulting in the very high Pearson $r$ value of $0.974$.

\end{document}